%% file: main.tex
\documentclass{article} 
\usepackage{iclr2024_conference,times}

\input{math_commands.tex}

\usepackage{enumitem}
\usepackage[hidelinks]{hyperref}
\usepackage{url}
\usepackage{booktabs}
\usepackage{amsfonts}
\usepackage{graphicx}
\usepackage{sidecap}
\usepackage[small]{caption}
\usepackage{subcaption}
\usepackage{amsmath}
\allowdisplaybreaks
\usepackage{amsthm}

\usepackage{wrapfig}

\usepackage{tabularx, booktabs}
\newcolumntype{C}{>{\centering\arraybackslash}X}
\newcolumntype{R}{>{\raggedleft\arraybackslash}X}
\newcolumntype{S}{>{\raggedleft\arraybackslash\hsize=.5\hsize}X}

\usepackage{arydshln}
\usepackage{booktabs}
\usepackage{multirow}

\usepackage{pifont}  %
\usepackage{makecell}
\newcommand{\es}[2]{#1_{#2}} %
\newcommand{\defn}[1]{\textbf{#1}}
\newcommand{\optparens}[1]{\if\relax\detokenize{#1}\relax\else(#1)\fi}   %
\newcommand{\inten}[2]{\lambda_{{#1}}\optparens{#2}} %

\usepackage[noabbrev,capitalize]{cleveref}
\crefname{equation}{equation}{equations}
\crefname{section}{section}{sections}
\crefname{footnote}{footnote}{footnotes}   
\crefname{line}{line}{lines}   
\crefname{assumption}{assumption}{assumptions}
\crefname{lstlisting}{listing}{listings}
\Crefname{lstlisting}{Listing}{Listings}
\usepackage{bbm}
\usepackage{appendix}

\definecolor{aigold}{RGB}{244,210, 1} 
\definecolor{aigreen}{RGB}{245, 255, 249}

\definecolor{humanpurple}{RGB}{235, 222, 240} 

\definecolor{commentgray}{RGB}{86, 101, 115}

\definecolor{aired}{RGB}{255,180,181}

\usepackage{listings}
\usepackage{parcolumns}
\lstdefinestyle{datalogstyle}{
	basicstyle={\codefont\small},  
	xleftmargin={6pt},
        xrightmargin={6pt},
        breakindent=0pt,
	frame=tb,
	stepnumber=1,
	firstnumber=1,
	numberfirstline=true,
	tabsize=2,
	showtabs=false,
	showspaces=false,
	showstringspaces=false,
	extendedchars=true,
	breaklines=true,
	columns=fullflexible,
	keepspaces=true,
	escapeinside={@}{@},
	firstnumber=last,
	captionpos=b,
	commentstyle=\color{black!65},
	numberstyle=\tiny\color{black!65},
	stringstyle=\color{codepurple},
	breakatwhitespace=false, 
	keepspaces=true,                 
	numbersep=5pt,                  
	showspaces=false,                
	showstringspaces=false,
	showtabs=false,
	aboveskip={0.8\baselineskip},
	belowskip={0.2\baselineskip},
	backgroundcolor=\color{aigreen},
}
\usepackage{soul}
\definecolor{rebuttal}{RGB}{229,255,204}
\sethlcolor{rebuttal}

\newcommand{\codefont}{\fontfamily{lmtt}\selectfont}

\title{EasyTPP: Towards Open Benchmarking\\Temporal Point Processes}

\author{Siqiao Xue$^{\diamondsuit}$, Xiaoming Shi$^{\diamondsuit}$, Zhixuan Chu$^{\diamondsuit}$, Yan Wang$^{\diamondsuit}$, Hongyan Hao$^{\diamondsuit}$, Fan Zhou$^{\diamondsuit}$, \\ 
\textbf{Caigao Jiang$^{\diamondsuit}$, Chen Pan$^{\diamondsuit}$, James Y. Zhang$^{\diamondsuit}$, Qingsong Wen$^{\spadesuit}$, Jun Zhou$^{\diamondsuit}$} \\
$^{\diamondsuit}$Ant Group, \quad $^{\spadesuit}$Alibaba Group \\
\texttt{\{siqiao.xsq, peter.sxm, chuzhixuan.czx\}@alibaba-inc.com} 
\And
Hongyuan Mei \\
TTIC \\
\texttt{hongyuan@ttic.edu} \\
}

\iclrfinalcopy 
\begin{document}

\maketitle

\begin{abstract}
Continuous-time event sequences play a vital role in real-world domains such as healthcare, finance, online shopping, social networks, and so on. To model such data, temporal point processes (TPPs) have emerged as the most natural and competitive models, making a significant impact in both academic and application communities. Despite the emergence of many powerful models in recent years, there hasn't been a central benchmark for these models and future research endeavors. This lack of standardization impedes researchers and practitioners from comparing methods and reproducing results, potentially slowing down progress in this field. 
In this paper, we present EasyTPP, the first central repository of research assets (e.g., data, models, evaluation programs, documentations) in the area of event sequence modeling. 
Our EasyTPP makes several unique contributions to this area: 
a unified interface of using existing datasets and adding new datasets; 
a wide range of evaluation programs that are easy to use and extend as well as facilitate reproducible research; 
implementations of popular neural TPPs, together with a rich library of modules by composing which one could quickly build complex models. 
All the data and implementation can be found at \href{https://github.com/ant-research/EasyTemporalPointProcess}{\textcolor{blue}{Github repository}}.
We will actively maintain this benchmark and welcome contributions from other researchers and practitioners. 
Our benchmark will help promote reproducible research in this field, thus accelerating research progress as well as making more significant real-world impacts.


\end{abstract}

\section{Introduction}

\begin{wrapfigure}{r}{0.6\textwidth}
    \includegraphics[width=0.6\textwidth]{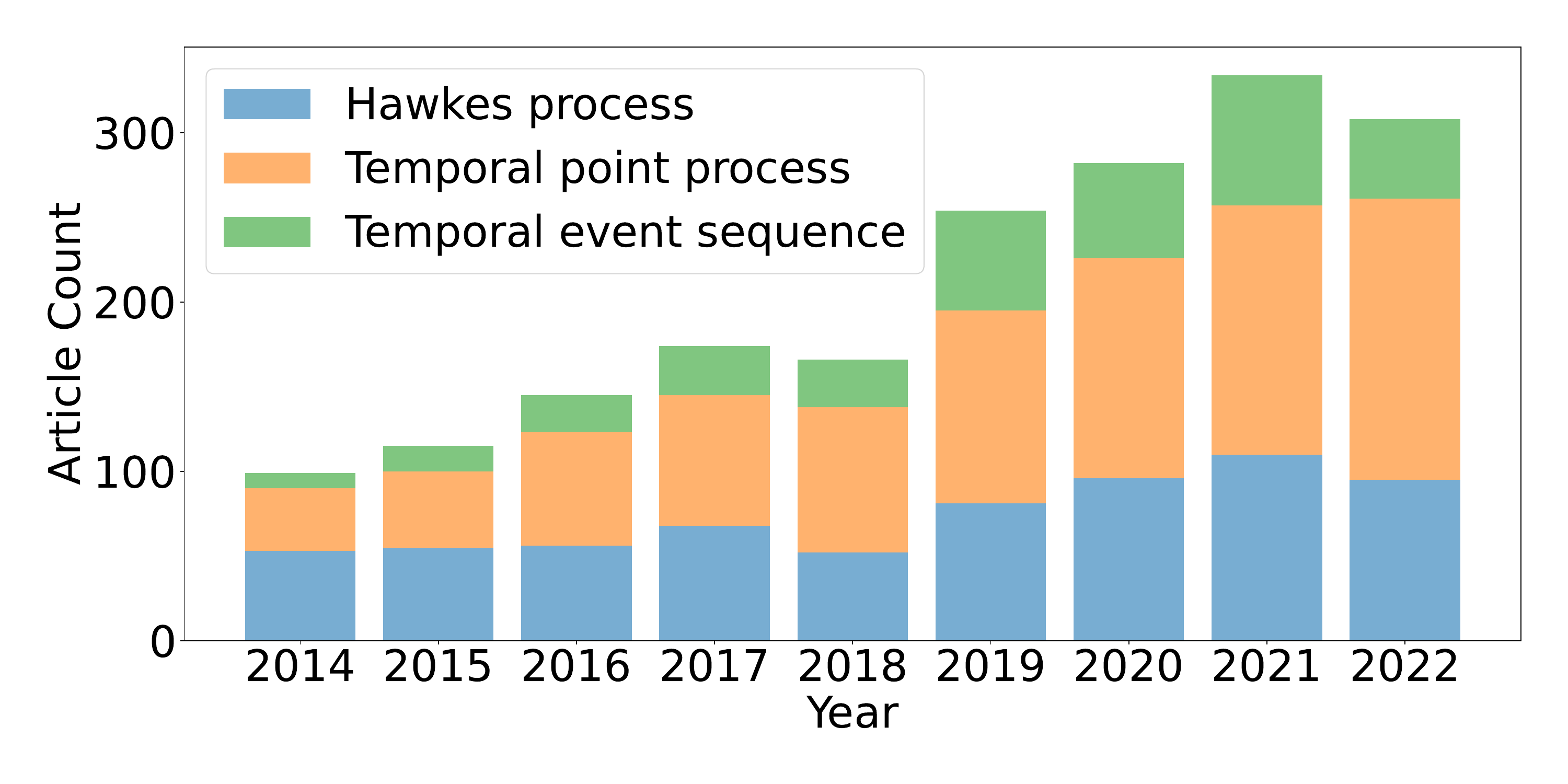}
  \caption{ArXiv submissions over time on TPPs. See \cref{app:cite} for details.}
  \label{fig:arxiv}
\end{wrapfigure}

Continuous-time event sequences are ubiquitous in various real-world domains, such as neural spike trains in neuroscience~\citep{williams2020point}, orders in financial transactions~\citep{jin2020visual}, and user page viewing behavior in the e-commerce platform~\citep{hernandez2017analysis}. To model these event sequences, temporal point processes (TPPs) are commonly used, which specify the probability of each event type's instantaneous occurrence, also known as the \emph{intensity function}, conditioned on the past event history. Classical TPPs, such as Poisson processes~\citep{daley-07-poisson} and Hawkes processes~\citep{hawkes-71}, have a well-established mathematical foundation and have been widely used to model traffic ~\citep{cramer1969historical}, finance~\citep{hasbrouck-1991} and seismology~\citep{ogata1988statistical} for several decades. However, the strong parametric assumptions in these models constrain their ability to capture the complexity of real-world phenomena. 


To overcome the limitations of classical TPPs, many researchers have been developing neural versions of TPPs, which leverage the expressiveness of neural networks to learn complex dependencies; see \cref{sec:related} for a comprehensive discussion. 
Since then, numerous advancements have been made in this field, as evidenced by the rapidly growing literature on neural TPPs since 2016. Recent reviews have documented the extensive methodological developments in TPPs, which have expanded their applicability to various real-world scenarios. As shown in \cref{fig:arxiv}~(see~\cref{app:cite}~for details of how we count the articles.), the number of research papers on TPPs has been steadily increasing, indicating the growing interest and potential impact of this research area. These advancements have enabled more accurate and flexible modeling of event sequences in diverse fields.

In this work, inspired by Hugging Face~\citep{wolf-etal-2020-transformers} for computer vision and natural language processing, we take the initiative to build a central library, namely EasyTPP,  of popular research assets (e.g., data, models, evaluation methods, documentations) with the following distinct merits:

\begin{enumerate}[leftmargin=*]
    \item \textbf{Standardization}. We establish a standardized benchmark to enable transparent comparison of models. 
    Our benchmark currently hosts 5 popularly-used real-world datasets that cover diverse real-world domains (e.g., commercial, social), and will include datasets in other domains (e.g., earthquake and volcano eruptions). 
    One of our contributions is to develop a unified format for these datasets and provide source code (with thorough documentation) for data processing. 
    This effort will free future researchers from large amounts of data-processing work, and facilitate exploration in new research topics such as transfer learning and adaptation (see \Cref{sec:future}). 
    \item \textbf{Comprehensiveness}. Our second contribution is to provide a wide range of easy-to-use evaluation programs, covering popular evaluation metrics (e.g., log-likelihood, kinds of next-event prediction accuracies and sequence similarities) and significance tests (e.g., permutation tests).
    By using this shared set of evaluation programs, researchers in this area will not only achieve a higher pace of development, but also ensure a better reproducibility of their results. 
    \item \textbf{Convenience}. Another contribution of EasyTPP is a rich suite of modules (functions and classes) which will significantly facilitate future method development. 
    We reproduced previous most-cited and competitive models by composing these modules like building LEGOs; other researchers can reuse the modules to build their new models, significantly accelerating their implementation and improving their development experience. Examples of modules are presented in \cref{sec:bench_process}. 
    \item \textbf{Flexibility}. Our library is compatible with both PyTorch~\citep{paszke2019pytorch} and TensorFlow~\citep{abadi2016tensorflow}, the top-2 popular deep learning frameworks, and thus offers a great flexibility for future research in method development.
    \item \textbf{Extensibility}. Following our documentation and protocols, one could easily extend the EasyTPP library by adding new datasets, new modules, new models, and new evaluation programs. 
    This high extensibility will contribute to building a healthy open-source community, eventually benefiting the research area of event sequence modeling. 
\end{enumerate}

\section{Background}
\paragraph{Definition.} Suppose we are given a fixed time interval $[0, T]$ over which an event sequence is observed. Suppose there are $I$ events in the sequence at times $0 < t_1 < \ldots < t_I \leq T$. We denote the sequence as $\es{x}{[0, T]} = (t_1, k_1), \ldots, (t_I, k_I)$ where each $k_i \in \{1, \ldots, K\}$ is a discrete event type. Note that representations in terms of time $t_i$ and the corresponding inter-event time $\tau_i=t_i-t_{i-1}$ are isomorphic, we use them interchangeably. TPPs are probabilistic models for such event sequences. If we use $p_k(t \mid \es{x}{[0,t)})$ to denote the probability that an event of type $k$ occurs over the infinitesimal interval $[t, t + dt)$, then the probability that nothing occurs will be $1 - \sum_{k=1}^{K} p_k(t \mid \es{x}{[0,t)})$. Formally, the distribution of a TPP can be characterized by the \defn{intensity} $\lambda_k( t \mid \es{x}{[0,t)}) \geq 0$ for each event type $k$ at each time $t > 0$ such that $p_k(t \mid \es{x}{[0,t)}) = \lambda_k( t \mid \es{x}{[0,t)}) dt$. 
\begin{figure}[t]
  \begin{center}
    \includegraphics[width=0.9\textwidth]{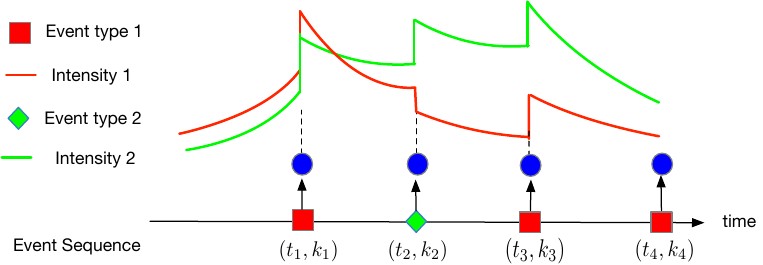}
  \end{center}
  \caption{Drawing an event stream from a neural TPP.  The model reads the sequence of past
events (polygons) to arrive at a hidden state (blue). That state determines the future "intensities" of the two types of events--that is, their time-varying instantaneous probabilities. The intensity functions are continuous
parametric curves (solid lines) determined by the most recent model state. 
Events will update the future intensity curves as they occur.}
  \label{fig:neural_tpp}
\end{figure}

\paragraph{Neural TPPs.} A neural TPP model autoregressively generates events one after another via neural networks. A schematic example is shown in \cref{fig:neural_tpp} and a detailed description on data samples can be found at the documentation of the repository. For the $i$-th event $(t_i, k_i)$, it computes the embedding of the event $\bm{e}_i \in \mathbb{R}^{D}$ via an embedding layer and the hidden state $\bm{h}_i$ gets updated conditioned on $\bm{e}_i$ and the previous state $\bm{h}_{i-1}$. Then one can draw the next event conditioned on the hidden state $\bm{h}_i$:
\begin{align}
    t_{i+1}, k_{i+1} \sim \mathbb{P}_{\theta}(t_{i+1}, k_{i+1} \vert \bm{h}_i), \quad \bm{h}_i = f_{update}(\bm{h}_{i-1}, \bm{e}_i),
    \label{eqn:recursion}
\end{align}
 where $f_{update}$ denotes a recurrent encoder, which could be either RNN~\citep{du-16-recurrent,mei-17-neuralhawkes} or more expressive attention-based recursion layer~\citep{zhang-2020-self,zuo2020transformer,yang-2022-transformer}. A new line of research models the evolution of the states completely in continuous time:
\begin{align}
    & \bm{h}_{i-} = f_{evo}(\bm{h}_{i-1}, t_{i-1}, t_i) \quad \text{between event times} \label{eqn:state_evo} \\
    & \bm{h}_i = f_{update}(\bm{h}_{i-}, \bm{e}_{i}) \quad \text{at event time $t_i$} 
\end{align}
The state evolution in \Cref{eqn:state_evo} is generally governed by an ordinary differential equation (ODE)~\citep{rubanova2019latent}. For a broad and fair comparison, in EasyTPP, we implement not only recurrent TPPs but also an ODE-based continuous-time state model.

\paragraph{Learning TPPs.}
Negative log-likelihood (NLL) is the default training objective for both classical and neural TPPs. The NLL of a TPP given the entire event sequence $\es{x}{[0, T]}$ is
\begin{align}
    \sum_{i=1}^{I} \log \lambda_{k_i}( t_i \mid \es{x}{[0,t_i)}) - \int_{t=0}^{T} \sum_{k=1}^{K} \lambda_{k}(t \mid \es{x}{[0,t)}) dt \label{eqn:autologlik}
\end{align}
Derivations of this formula can be found in previous work~\citep{hawkes-71,mei-17-neuralhawkes}.

\section{The Benchmarking Pipeline}
\label{sec:bench_process}
\begin{figure}
  \begin{center}
    \includegraphics[width=\textwidth]{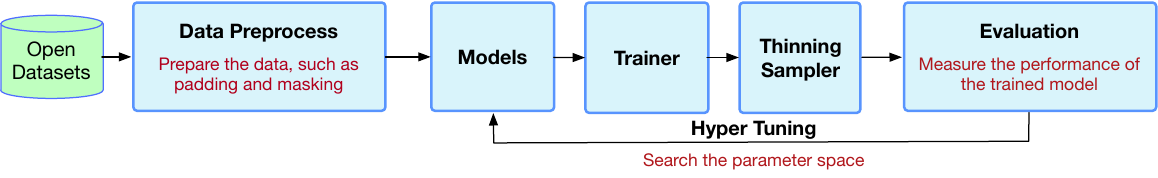}
  \end{center}
  \caption{An open benchmarking pipeline using EasyTPP.}
  \label{fig:pipeline}
\end{figure}

\cref{fig:pipeline} presents the open benchmarking pipeline implemented in EasyTPP. 
This pipeline will facilitate future research in this area and help promote reproducible work. 
In this section, we introduce and discuss each of the key components.

\paragraph{Data Preprocessing.}  Following common practices, we split the set of sequences into disjoint train, validation, and test set. Normally, the input is fed batch-wise into the model; there may exist sequences of events that have unequal-length in the same batch. To feed the sequences of varying lengths into the model, we pad all sequences to the same length, then use the ``sequence$\_$mask'' tensor to identify which event tokens are padding. As we implemented several variants of attention-based TPPs, we also generated the ``attention$\_$mask'' to mask all the future positions at each event to avoid ``peeking into the future''. The padding and masking mechanism is the same as that used in NLP filed. See \cref{app:padding} for a detailed explanation on sequence padding and masking in EasyTPP.

\paragraph{Model Implementation.} 
Our EasyTPP library provides a suite of modules, and one could easily build complex models by composing these modules. Specifically, we implemented the models (see \cref{sec:setup}) evaluated in this paper with our suite of modules (e.g., continuous-time LSTM, continuous-time attention). 
Moreover, some modules are model-agnostic methods for training and inference, which will further speed up the development speed of future methodology research. 
Below are two signature examples: 
\begin{itemize}[leftmargin=*]
    \item compute$\_$loglikelihood (function), which calculates log-likelihood of a model given data. It is non-trivial to correctly implement it due to the integral term of log-likelihood in \Cref{eqn:autologlik}, and we have found errors in popular implementations. 
    \item EventSampler (class), which draws events from a given point process via the thinning algorithm. The thinning algorithm is commonly used in inference but it is non-trivial to implement (and rare to see) an efficient and batched version. Our efficient and batched version (which we took great efforts to implement) will be useful for nearly all intensity-based event sequence models. 
\end{itemize}

\paragraph{Training.}  We can estimate the model parameters by locally maximizing the NLL in \Cref{eqn:autologlik} with any stochastic gradient method. Note that computing the NLL can be challenging due to the presence of the integral in the second term in \Cref{eqn:autologlik}. In EasyTPP, by default, we approximate the integral by Monte-Carlo estimation to compute the overall NLL (see \cref{app:loglike}). Nonetheless, EasyTPP also incorporates intensity-free models~\citep{shchur-20-intensity}, whose objectives are easier to compute.

\paragraph{Sampling.}  Given the learned parameters, we apply the minimum Bayes risk (MBR) principle to predict the time and type with the lowest expected loss. A recipe can be found in \cref{app:prediction}. Note that other methods exist for predicting with a TPP, such as adding an MLP layer to directly output the time and type prediction~\citep{zuo2020transformer,zhang-2020-self}. 
However, the primary focus of this benchmark is generative models of event sequences, for which the principal approach is the thinning algorithm ~\citep{ogata1988statistical}. 
In EasyTPP, we implemented a batched version of thinning algorithm, which is then used to evaluate the TPPs in our experiments.

\paragraph{Hyperparameter Tuning.} 
Our EasyTPP benchmark provides programs for automatic hyperparameter tuning. 
In addition to classical grid search, we also integrate \emph{Optuna}~\citep{optuna_2019} in our framework to adaptively prune the search grid.

\section{Software Interface}
\label{section:interface}

The purpose of building EasyTPP is to provide a simple and standardized framework to allow users to apply different state-of-the-art (SOTA) TPPs to any datasets as they would like to. 
For researchers, EasyTPP provides an implementation interface to integrate new recourse methods
in an easy-to-use way, which allows them to compare their method to already existing methods. See the pseudo implementation in \cref{lst:implement_model}. For industrial practitioners, the availability of benchmarking code helps them easily assess the applicability of TPP models for their own problems. See an example of running an existed model in EasyTPP in \cref{lst:run_model}. The full documentation of software interfaces can be found at our \href{https://github.com/ant-research/EasyTemporalPointProcess/tree/main/docs/source/ref}{\textcolor{blue}{repo}}. For more details of software architecture, please see \href{https://github.com/ant-research/EasyTemporalPointProcess/blob/main/docs/source/get_started/introduction.rst}{\textcolor{blue}{introduction part of online documentation}} and \cref{app:interface} as well.

\section{Experimental Evaluation}\label{sec:exp}

\subsection{Experimental Setup}\label{sec:setup}

We comprehensively evaluate 9 models in our benchmark, which include the classical \textbf{Multivariate Hawkes Process (MHP)} with an exponential kernel, (see\cref{app:model_implement} for more details), and 8 widely-cited state-of-the-art neural models:
\begin{itemize}[leftmargin=*]
    \item Two RNN-based models: \textbf{Recurrent marked temporal point process (RMTPP)} \citep{du-16-recurrent} and \textbf{neural Hawkes Process (NHP)} \citep{mei-17-neuralhawkes}.
    \item Three attention-based models: \textbf{self-attentive Hawkes pocess (SAHP)} \citep{zhang-2020-self}, \textbf{transformer Hawkes process (THP)} \citep{zuo2020transformer}, \textbf{attentive neural Hawkes process (AttNHP)} \citep{yang-2022-transformer}.
    \item One TPP with the fully neural network based intensity: \textbf{FullyNN} \citep{omi-19-fully}.
    \item One intensity-free model \textbf{IFTPP} \citep{shchur-20-intensity}.
    \item One TPP with the hidden state evolution governed by a neural ODE: \textbf{ODETPP}. It is a simplified version of the TPP proposed by \citet{chen2021neural} by removing the spatial component. 
    \textbf{}.
\end{itemize}

\begin{minipage}[t]{0.5\linewidth}
\begin{lstlisting}[caption={Pseudo implementation of customizing a TPP model in PyTorch using EasyTPP.},label={lst:implement_model}]
from easy_tpp.model.torch_model.torch_basemodel import TorchBaseModel

# Custom TPP implementations need to inherit from the BaseModel interface
class NewModel(TorchBaseModel):
    def __init__(self, model_config):
        super(NewModel, self).__init__(model_config)

    # Forward along the sequence, output the states / intensities at event times
    def forward(self, batch):
        @
\vdots
@
        return states

    # Compute the loglikelihood loss
    def loglike_loss(self, batch):
        @
        \vdots 
        @
        return loglike

    # Compute the intensities at given sampling times, used by Thinning sampler
    def compute_intensities_at_sample_times(self, batch, sample_times, **kwargs):
        @
        \vdots 
        @
        return intensities
\end{lstlisting}
\end{minipage}
\begin{minipage}[t]{0.5\linewidth}
\begin{lstlisting}[caption={Example implementation of running a TPP model using EasyTPP.},label={lst:run_model}]
import argparse
from easy_tpp.config_factory import Config
from easy_tpp.runner import Runner

def main():
    parser = argparse.ArgumentParser()
    parser.add_argument('--config_dir', 
                        type=str, 
                        required=True, 
                        help='Dir of config to train and evaluate the model.')
    parser.add_argument('--experiment_id', 
                        type=str, 
                        required=True, 
                        help='Experiment id in the config file.')

    args = parser.parse_args()

    # Build up the configuation 
    config = Config.build_from_yaml_file(args.config_dir, args.experiment_id)

    # Intialize the runner of pipeline
    model_runner = Runner.build_from_config(config)

    # Start running
    model_runner.run()

if __name__ == '__main__':
    main()
\end{lstlisting}
\end{minipage}

We conduct experiments on $1$ synthetic and $5$ real-world datasets from popular works that contain diverse characteristics in terms of their application domains and temporal statistics (see \Cref{tab:stats_dataset}): 
\begin{itemize}[leftmargin=*]
    \item \textbf{Synthetic}. This dataset contains synthetic event sequences from a univariate Hawkes process sampled using {Tick} \citep{tick-2017} whose conditional intensity function is defined by
$\lambda(t) = \mu +  \sum_{t_{i} < t} \alpha \beta \cdot \text{exp}\left({-\beta(t - t_{i})}\right)$
with $\mu = 0.2,  \alpha = 0.8,  \beta = 1.0$. We randomly sampled disjoint train, dev, and test sets with $1200$, $200$ and $400$ sequences.

      \item {\textbf{Amazon}}\citep{amazon-2018}. This dataset includes time-stamped user product reviews behavior from January, 2008 to October, 2018. 
Each user has a sequence of produce review events with each event containing the timestamp and category of the reviewed product, with each category corresponding to an event type. We work on a subset of $5200$ most active users with an average sequence length of $70$ and then end up with $K=16$ event types. 

    \item \textbf{Retweet} 
    \textnormal{\citep{zhou-2013}}. This dataset contains time-stamped user retweet event sequences.  The events are categorized into $K = 3$ types: retweets by “small,” “medium” and “large” users. Small users have fewer than $120$ followers, medium users have fewer than $1363$, and the rest are large users. We work on a subset of $5200$ active users with an average sequence length of $70$.

    \item \textbf{Taxi} \textnormal{\citep{whong-14-taxi}}. This dataset tracks the time-stamped taxi pick-up and drop-off events across the five boroughs of the New York City; each (borough, pick-up or drop-off) combination defines an event type, so there are $K=10$ event types in total. We work on a randomly sampled subset of $2000$ drivers with an average sequence length of $39$.

    \item \textbf{Taobao} \citep{xue2022hypro}. This dataset contains time-stamped user click behaviors on Taobao shopping pages from November 25 to December 03, 2017. Each user has a sequence of item click events with each event containing the timestamp and the category of the item. The categories of all items are first ranked by frequencies and the top $19$ are kept while the rest are merged into one category, with each category corresponding to an event type. We work on a subset of $4800$ most active users with an average sequence length of $150$ and then end up with $K=20$ event types.

    \item {\textbf{StackOverflow} \textnormal{\citep{snapnets}}.} This dataset has two years of user awards on a question-answering website: each user received a sequence of badges and there are $K=22$ different kinds of badges in total.  We work on a subset of $2200$ active users with an average sequence length of $65$.
\end{itemize}



\noindent \textbf{Evaluation Protocol.} We keep the model architectures as the original implementations in their papers. For a fair comparison, we use the same training procedure for all the models: we used Adam \citep{kingma-15} with the default parameters, biases initialized with
zeros, no learning rate decay, the same maximum number of training epochs, and early stopping criterion (based on log-likelihood on the held-out dev set) for all models.

We mainly examine the models in two standard scenarios. 
\begin{itemize}[leftmargin=*]
    \item Goodness-of-fit: we fit the models on the train set and measure the log-probability they assign to the held-out data.
    \item Next-event prediction: we use the minimum Bayes risk (MBR) principle to predict the next event time given only the preceding events, as well as its type given both its true time and the preceding events. We evaluate the time and type prediction by RMSE and error rate, respectively.
\end{itemize}
\begin{figure*}[t]
    \begin{center}
        \begin{subfigure}[t]{0.3\linewidth}
                \includegraphics[width=0.99\linewidth]{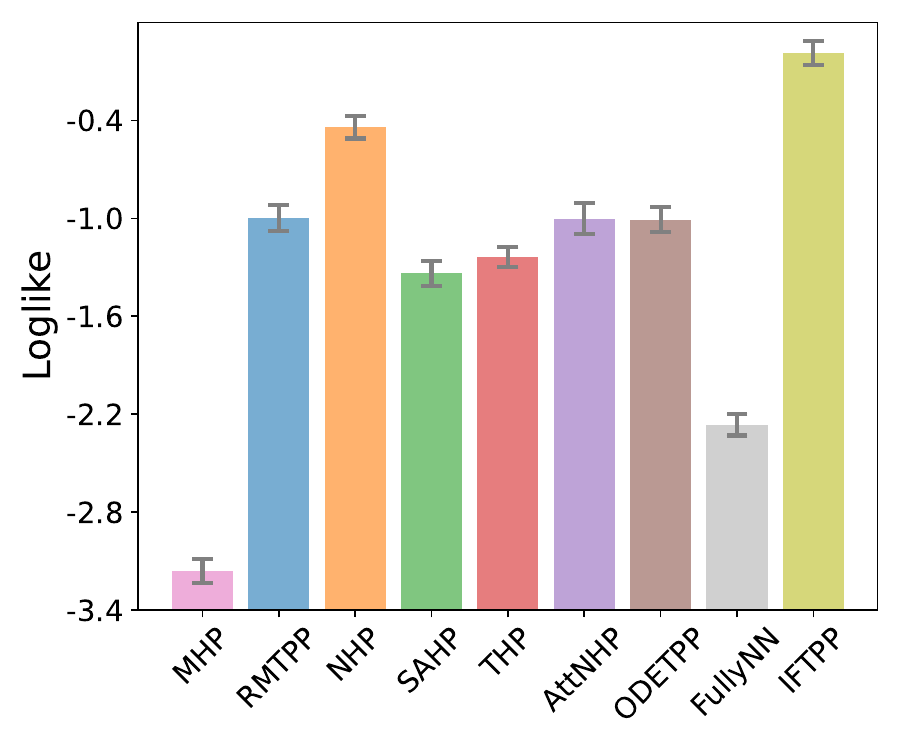}
			\caption{Synthetic 1-D Hawkes}
        \end{subfigure}
        \hfill
        \begin{subfigure}[t]{0.3\linewidth}
        \includegraphics[width=0.99\linewidth]{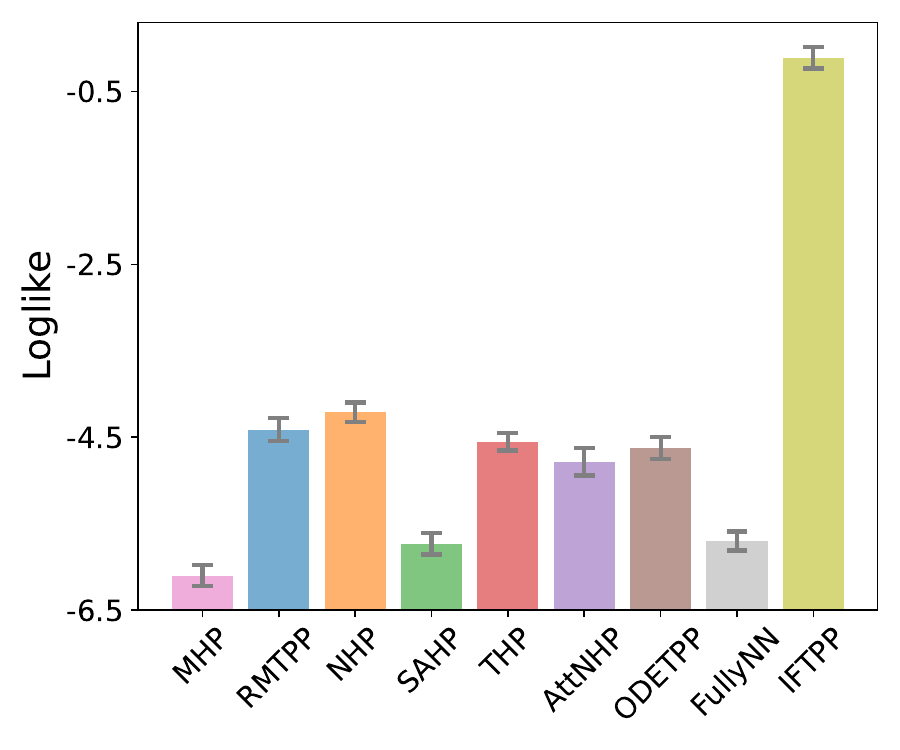}
			\caption{Retweet}
                \label{fig:loglike_retweet}
        \end{subfigure}
        \hfill
        \begin{subfigure}[t]{0.3\linewidth}
        \includegraphics[width=0.99\linewidth]{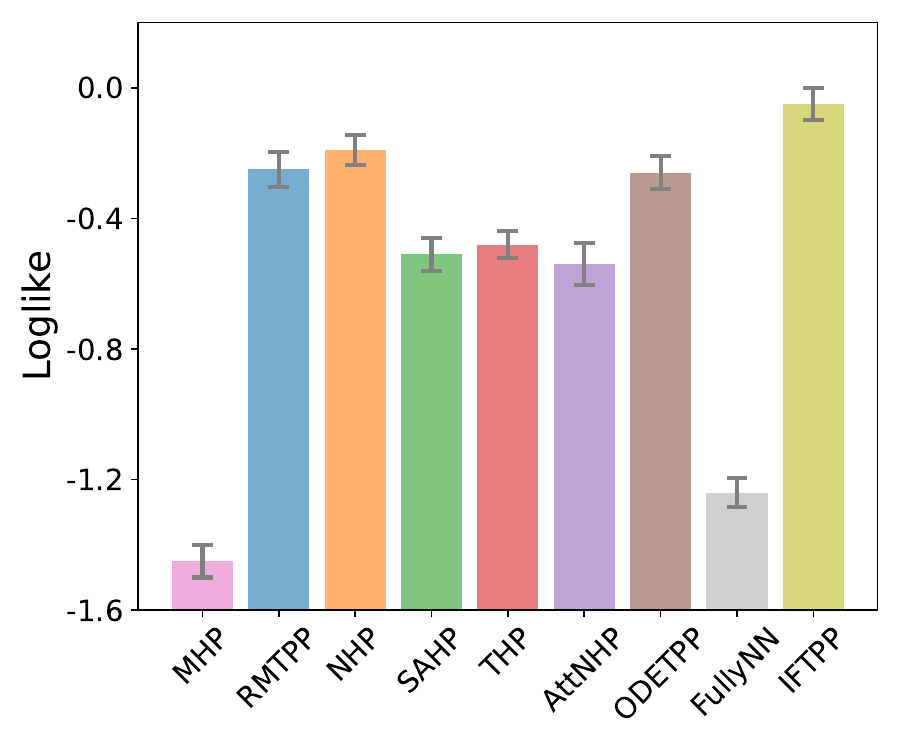}
                \caption{Taxi}
                \label{fig:loglike_taxi}
        \end{subfigure}
    \caption{Performance of all the methods on the goodness-of-fit task on synthetic Hawkes, Retweet, and Taxi data. A higher score is better. All methods are implemented in PyTorch.}\label{fig:main_results_loglike}
	\end{center}
	\vspace{-4pt}
\end{figure*}

In addition, we propose a new evaluation task: the long-horizon prediction. Given the prefix of each held-out sequence $x_{[0, T]}$, we autoregressively predict the next events in a future horizon $\hat{x}_{(T, T^{'}]}$. It is evaluated by measuring the optimal transport distance (OTD), a type of edit distance for event sequences~\citep{mei-19-smoothing}, between the prediction $\hat{x}_{(T, T^{'}]}$ and ground truth $x_{(T, T^{'}]}$. As pointed out by~\citet{xue2022hypro}, long-horizon prediction of event sequences is essential in various real-world domains, and this task provides new insight into the predictive performance of the models.

It is worth noting that the original version of the FullyNN model does not support multi-type event sequences. Therefore it is excluded from the type prediction task.

\begin{table*}[tbh]
	\begin{center}
		\begin{small}
			\begin{sc}
				\begin{tabularx}{1.00\textwidth}{l *{1}{S}*{5}{R}}
					\toprule
					Model & \multicolumn{5}{c}{Metrics (Time RMSE $/$ Type Error Rate)}  \\
					\cmidrule(lr){2-6}
					& Amazon & Retweet & Taxi & Taobao & StackOverflow   \\
					\midrule
MHP	& $0.635/75.9\%$ & $22.92/55.7\%$ & $0.382/9.53\%$ & $0.539/68.1\%$ & $1.388/65.0\%$  \\	
 & $\small 0.005/0.005$ & $0.212/0.004$ & $0.002/0.0004$ & $0.004/0.004$ & $0.011/0.005$  \\		\midrule
 RMTPP	& $0.620/68.1\%$ & $22.31/44.1\%$ & $0.371/9.51\%$ & $0.531/55.8\%$ & $1.376/57.3\%$  \\	
 & $\small 0.005/0.006$ & $0.209/0.003$ & $0.003/0.0003$ & $0.005/0.004$ & $0.018/0.005$  \\
 \midrule
        NHP & $0.621/67.1\%$ & $\underline{21.90}/40.0\%$ & $\underline{0.369}/\underline{8.50\%}$ & $0.531/54.2\%$ & $\underline{1.372}/\underline{55.0\%}$ \\
        & $\small 0.005/0.006$ & $0.184/0.002$ & $0.003/0.0005$ & $0.005/0.006$ & $0.011/0.006$  \\
        \midrule
        SAHP & $0.619/67.7\%$ & $22.40/41.6\%$ & $0.372/9.75\%$ & $0.532/54.6\%$ & $1.375/56.1\%$ \\
        & $\small 0.005/0.006$ & $0.301/0.002$ & $0.003/0.0008$ & $0.004/0.002$ & $0.013/0.005$  \\
        \midrule
        THP & $0.621/66.1\%$ & $22.01/41.5\%$ & $0.370/8.68\%$ & $0.531/\underline{53.6\%}$ & $1.374/55.0\%$ \\
        & $\small 0.003/0.007$ & $0.188/0.003$ & $0.003/0.0006$ & $0.003/0.004$ & $0.021/0.006$  \\
        \midrule
        AttNHP & $0.621/\underline{65.3\%}$ & $22.19/40.1\%$ & $0.371/8.71\%$ & $\underline{0.529}/53.7\%$ & $\underline{1.372}/55.2\%$ \\
        & $\small 0.005/0.006$ & $0.180/0.003$ & $0.003/0.0004$ & $0.005/0.001$ & $0.019/0.003$  \\
        \midrule
        ODETPP & $0.620/65.8\%$ & $22.48/43.2\%$ & $0.371/10.54\%$ & $0.533/55.4\%$ & $1.374/56.8\%$ \\
        & $\small 0.006/0.008$ & $0.175/0.004$ & $0.003/0.0008$ & $0.005/0.007$ & $0.022/0.004$  \\
        \midrule
        FullyNN & $\underline{0.615}/$N.A. & $21.92/$N.A.\phantom{x} & $0.373/$N.A.\phantom{x} & $0.529/$N.A.\phantom{x} & $1.375/$N.A.\phantom{x} \\
        & $\small 0.005/$N.A. & $0.159/$N.A.\phantom{x} & $0.003/$N.A.\phantom{x} & $0.005$N.A.\phantom{x} & $0.015/$N.A.\phantom{x}  \\
        \midrule
        IFTPP & $0.618/67.5\%$ & $22.18/\underline{39.7\%}$ & $0.377/8.56\%$ & $0.531/55.4\%$ & $1.373/55.1\%$ \\
        & $\small 0.005/0.007$ & $0.204/0.003$ & $0.003/0.006$ & $0.005/0.004$ & $0.010/0.005$  \\
					\bottomrule
				\end{tabularx}
			\end{sc}
		\end{small}
	\end{center}
	\caption{Performance of all the methods on next-event's time prediction and next-event's type prediction on five real datasets (for each model, first row corresponds to the metrics value while second row corresponds to the standard deviation). Lower score is better. All methods are implemented in \textcolor{blue}{PyTorch}. As clarified, FullyNN is not applicable for the type prediction tasks.}
	\label{tab:main_results_pred_main_text}
\end{table*}

\subsection{Results and Analysis}
\cref{fig:main_results_loglike} (see \cref{tab:main_results_numbers_loglike} for exact numbers in the figure) reports the log-likelihood on three held-out datasets for all the methods. We find IFTPP outperforms all the competitors because it evaluates the log-likelihood in a close form while the others (RMTPP, NHP, THP, AttNHP, ODETPP) compute the intensity function via Monte Carlo integration, causing numerical approximation errors. FullyNN method, which also exactly computes the log-likelihood, has worse fitness than other neural competitors. As \citet{shchur-20-intensity} points out, the PDF of FullyNN does not integrate to $1$ due to a suboptimal choice of the network architecture, therefore causing a negative impact on the performance. 
\begin{figure}
\begin{minipage}[t]{0.49\linewidth}
    \includegraphics[width=0.48\linewidth]{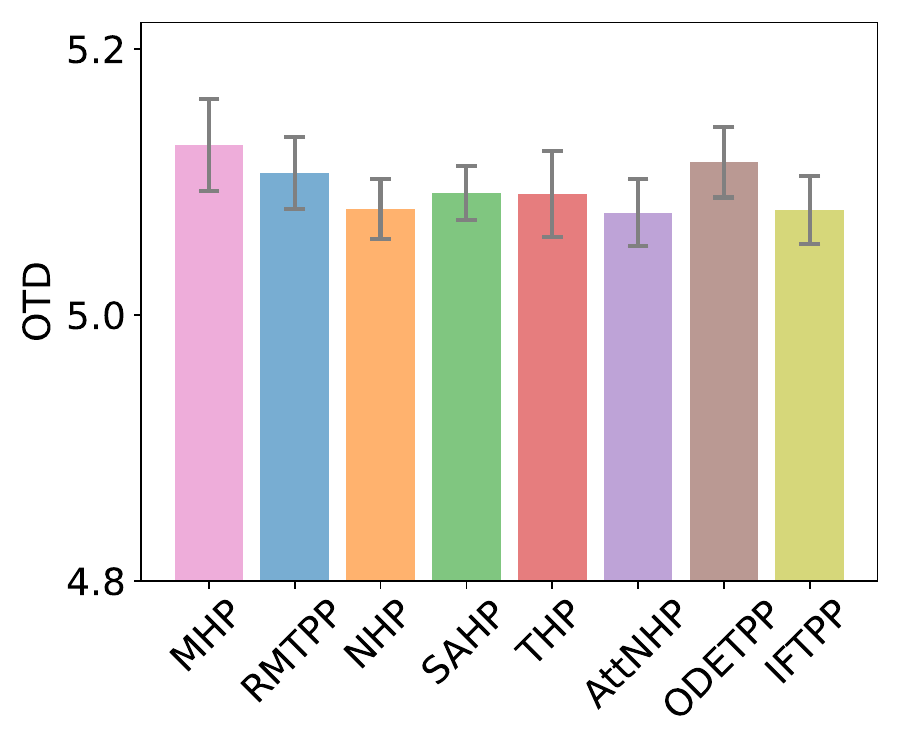}
    ~
    \includegraphics[width=0.48\linewidth]{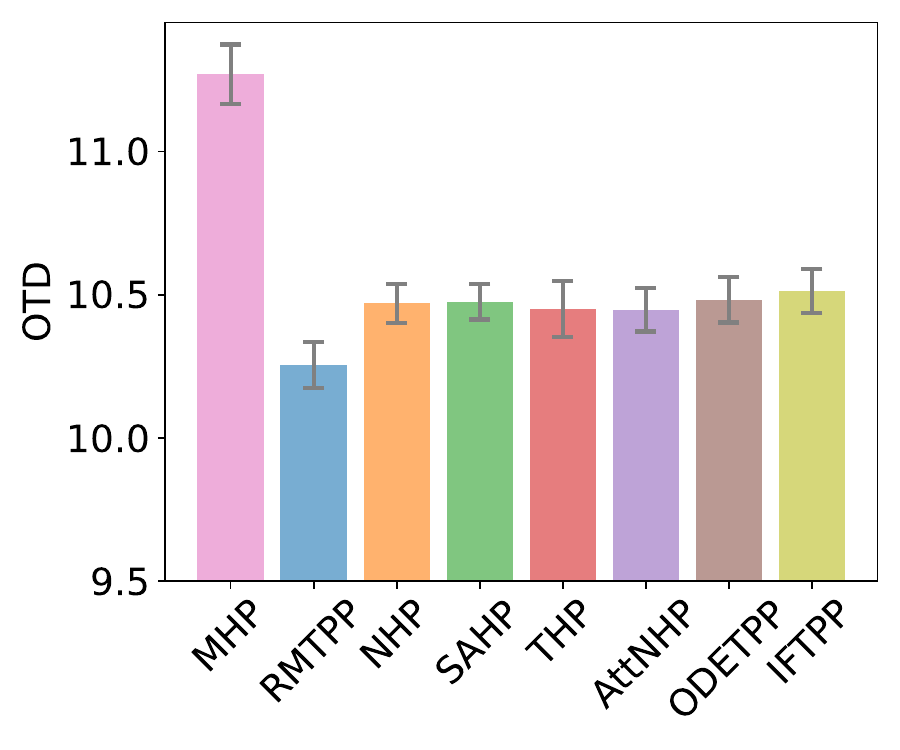}
    \caption{Long horizon prediction on Retweet data: left (avg prediction horizon $5$ events) vs. right (avg prediction horizon $10$ events).}
    \label{fig:main_results_retweet_lt_generation}
\end{minipage}
\hfill
\begin{minipage}[t]{0.49\linewidth}
    \includegraphics[width=0.48\linewidth]{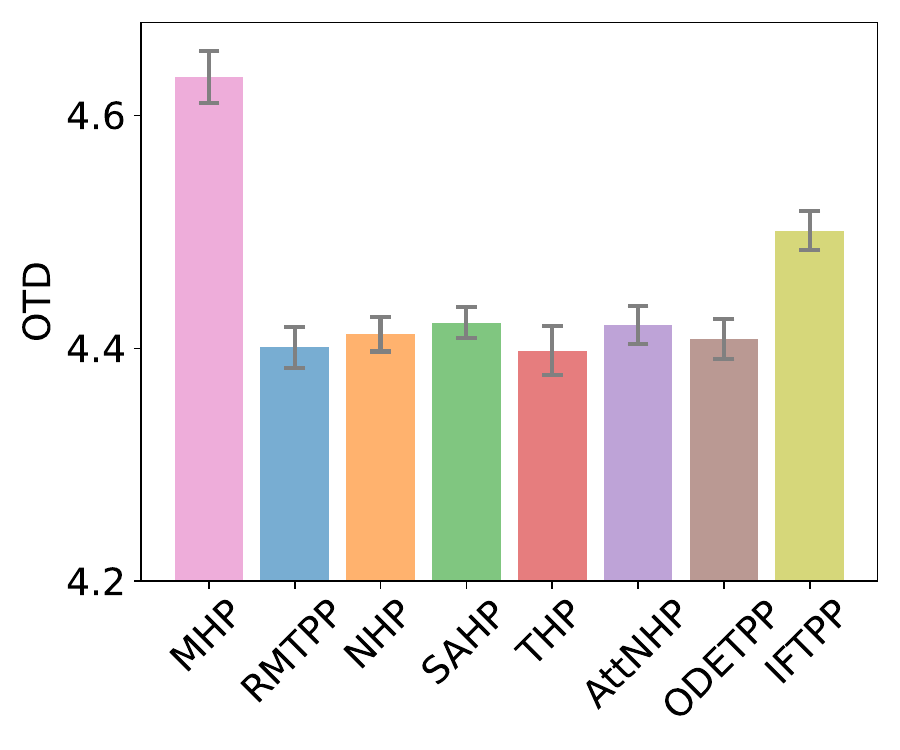}
    ~
    \includegraphics[width=0.48\linewidth]{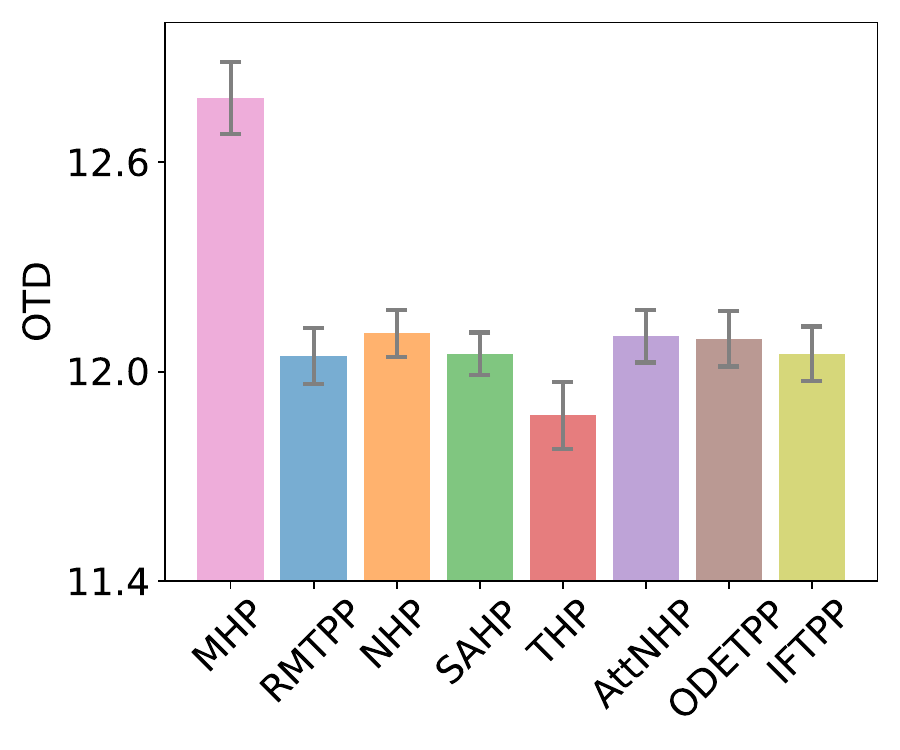}
    \caption{Long horizon prediction on Taxi data: left (avg prediction horizon $5$ events) vs. right (avg prediction horizon $10$ events).}
    \label{fig:main_results_taxi_lt_generation}
\end{minipage}
\vspace{-10pt}
\end{figure}

\cref{tab:main_results_pred_main_text} reports the time and type prediction results on five real datasets. We find there is no single winner against all the other methods. Attention-based methods (SAHP, THP, AttNHP) generally perform better than or close to non-attention methods (RMTPP, NHP, ODETPP,FullyNN and IFTPP) on Amazon, Taobao, and Stackoverflow, while NHP is the winner on both Retweet and Taxi. We see that NHP is a comparably strong baseline with attention-based TPPs. This is not too surprising because similar results have been reported in previous studies~\citep{yang-2022-transformer}.   

Not surprisingly, the performance of the classical model MHP is worse than the neural models across most of the evaluation tasks, consistent with the previous findings that neural TPPs have demonstrated to be more effective than classical models at fitting data and making predictions.
    
Please see \cref{app:more_result} for more results (e.g., actual numbers of figures) on all the datasets. With a growing number of TPP methods proposed, we will continuously expand the catalog of models and datasets and actively update the benchmark in our repository.


\paragraph{Analysis-I: Long Horizon Prediction.} We evaluate the long horizon prediction task on Retweet and Taxi datasets. On both datasets, we set the prediction horizon to be the one that approximately has $5$ and $10$ events, respectively. Shown in \cref{fig:main_results_retweet_lt_generation} and \cref{fig:main_results_taxi_lt_generation}, we find that AttNHP and THP are two co-winners on Retweet and THP is a single winner on Taxi. Nonetheless, the margin of the winner over the competitors is small. The exact numbers shown in these two figures could be found in \Cref{tab:main_results_retweet_lt_generation} in \Cref{app:more_result}. Because these models are autoregressive and locally normalized, they are all exposed to cascading errors. To fix this issue, one could resort to other kinds of models such as a hybridly normalized model~\citep{xue2022hypro}, which is out of the scope of the paper.

\paragraph{Analysis-II: Models with Different Frameworks---PyTorch vs.\@ TensorFlow.} Researchers normally implement their experiments
and models for specific ML frameworks. For example, recently proposed methods are mostly
restricted to PyTorch and are not applicable to TensorFlow models. As explained in \Cref{section:interface}, to facilitate the use of TPPs, we implement two equivalent sets of methods in PyTorch and TensorFlow. \cref{tab:main_results_rel_diff} in \cref{app:more_result} shows the relative difference between the results of Torch and TensorFlow implementations are all within $[-1.5\%, 1.5\%]$. To conclude, the two sets of models produce similar performance in terms of predictive ability. 

\section{Future Research Opportunities}
\label{sec:future}
We summarize our thoughts on future research opportunities inspired by our benchmarking results.

Most importantly, the results seem to be signaling that we should think beyond architectural design. For the past decade, this area has been focusing on developing new architectures, but the performance of new models on the standard datasets seem to be saturating. Notably, all the best to-date models make poor predictions on time of future events. Moreover, on type prediction, attention-based model~\citep{zuo2020transformer,zhang-2020-self,yang-2022-transformer} only outperform other architectures by a small margin. Looking into the future, we advocate for a few new research directions that may bring significant contributions to the field.

The first is to build foundation models for event sequence modeling. The previous model-building work all learns data-specific weights, and does not test the transferring capabilities of the learned models. Inspired by the emergence of foundation models in other research areas, we think it will be beneficial to explore the possibility to build foundation models for event sequences. Conceptually, learning from a large corpus of diverse datasets---like how GPTs~\citep{nakano2021webgpt} learn by reading open web text---has great potential to improve the model performance and generalization beyond what could be achieved in the current in-domain in-data learning paradigm. Our library can facilitate exploration in this direction since we unify the data formats and provide an easy-to-use interface that users can seamlessly plug and play any set of datasets. Challenges in this direction arise as different datasets tend to have disjoint sets of event types and different scales of time units.

The second is to go beyond event data itself and utilize external sources to enhance event sequence modeling. Seeing the performance saturation of the models, we are inspired to think whether the performance has been bounded by the intrinsic signal-to-noise ratio of the event sequence data. Therefore, it seems natural and beneficial to explore the utilization of other information sources, which include but are not limited to: (i) sensor data such as satellite images and radiosondes signals; (ii) structured and unstructured knowledge bases (e.g., databases, Wikipedia); (iii) large pretrained models such as ChatGPT~\citep{brown-2020-gpt} and GPT-4~\citep{gpt4}, whose rich knowledge and strong reasoning capabilities may assist event sequence models in improving their prediction accuracies. 
Concurrent with this work, \citet{shi2023language} has made an early step in this direction. 

The third is to go beyond observational data and embed event sequence models into real-world interventions~\citep{qu-2022-rltpp}. With interventional feedback from the real world, an event sequence model would have the potential to learn real causal dynamics of the world, which may significantly improve prediction accuracy.

All the aforementioned directions open up research opportunities for technical innovations.

\section{Related work}\label{sec:related}
\paragraph{Temporal Point Processes.} Over recent years, a large variety of RNN-based TPPs have been proposed \citep{du-16-recurrent,mei-17-neuralhawkes,xiao-17-modeling,omi-19-fully,shchur-20-intensity,mei-2020-datalog,boyd-20-vae}. Models of this kind enjoy continuous state spaces and flexible transition functions, thus achieving superior performance on many real datasets, compared to the classical Hawkes process~\citep{hawkes-71}. To properly capture the long-range dependency in the sequence, the attention and transformer techniques~\citep{vaswani-2017-transformer} have been adapted to TPPs~\citep{zuo2020transformer,zhang-2020-self,yang-2022-transformer,wen2022transformers} and makes further improvements on predictive performance. 
There has also been research in creative ways of training temporal point processes, such as in a meta learning framework~\citep{bae2023meta}. 
Despite significant progress made in academia, the existing studies usually perform model evaluations and comparisons in an ad-hoc manner, e.g., by using different experimental settings or different ML frameworks. Such conventions not only increase the difficulty in reproducing
these methods but also may lead to inconsistent experimental results among them.


\paragraph{Open Benchmarking on TPPs.} The significant attention attracted by TPPs in recent years naturally leads to a high demand for an open benchmark to fairly compare against baseline models. While many efforts have been made in the domains of recommender systems~\citep{fuxi-2021} and natural language processing \citep{glue-2019}, benchmarking TPPs is an under-explored topic. {Tick}~\citep{tick-2017} and {pyhawkes}\footnote{\small\url{https://github.com/slinderman/pyhawkes}} are two well-known libraries that focus on statistical learning for classical TPPs, which are not suitable for the SOTA neural models. {Poppy}~\citep{xu2018poppy} is a PyTorch-based toolbox for neural TPPs, but it has not been actively maintained since 2021 and has not implemented any recent SOTA methods. To the best of our knowledge, EasyTPP is the first package that provides open benchmarking for popular neural TPPs.

\section{Conclusion}
In this work, we presented EasyTPP, an open and comprehensive
benchmark for standardized and transparent comparison of TPP models. 
The benchmark hosts a diversity of datasets and models. 
In addition, it provides a user-friendly interface and a rich library, with which one could easily integrate new datasets and implement new models. 
With these features, EasyTPP has the potential to significantly facilitate future research in the area of event sequence modeling.

\bibliography{main}
\bibliographystyle{icml2020_url}

\clearpage
\appendix
\appendixpage

\section{{EasyTPP}'s Software Interface Details}
\label{app:interface}
In this section, we describe the architecture of our open-source benchmarking software {EasyTPP} in more detail and provide examples of different use cases and their implementation.

\subsection{High Level Software Architecture}
The purpose of building {EasyTPP} is to provide a simple and standardized framework to allow users to apply different state-of-the-art (SOTA) TPPs to arbitrary data sets. For researchers, {EasyTPP} provides an implementation interface to integrate new recourse methods
in an easy-to-use way, which allows them to compare their method to already existing methods. For industrial practitioners, the availability of benchmarking code helps them easily assess the applicability of TPP models for their own problems. 

\begin{figure}[t]
  \begin{center}
    \includegraphics[width=\textwidth]{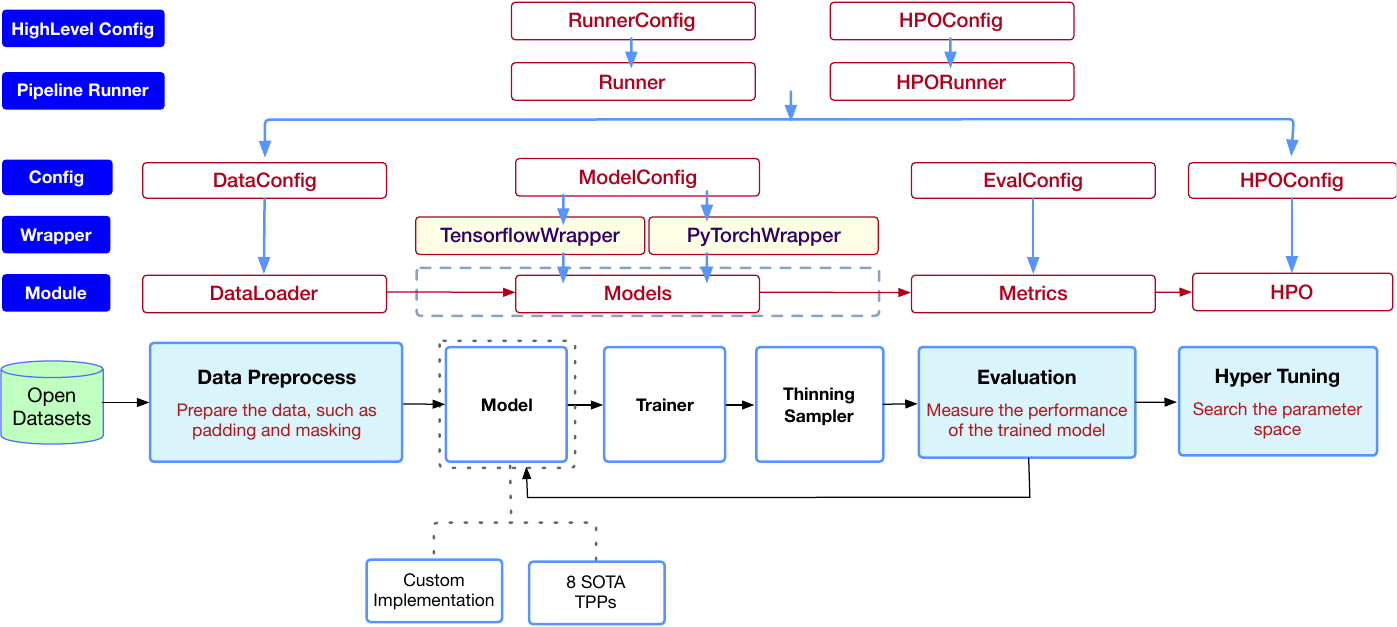}
  \end{center}
  \caption{Architecture of the EasyTPP library. The dashed arrows show the different implementation possibilities, either to use pre-defined SOTA TPP models or provide a custom implementation. All dependencies between the configurations and modules are visualized by solid arrows with
additional descriptions. 
    }
  \label{fig:pipeline_highlevel}
\end{figure}

A high level visualization of the {EasyTPP}'s software architecture is depicted in \cref{fig:pipeline_highlevel}. \emph{Data Preprocess} component provides a common way to access the event data across the software and maintains information about the
features. For the \emph{Model} component, the library provides the possibility to use existing methods or extend the users' custom methods and implementations. A \emph{wrapper} encapsulates the black-box models along with the trainer and sampler. The primary purpose of the wrapper is to provide a common
interface to easily fit in the training and evaluation pipeline, independently of their framework (e.g., PyTorch, TensorFlow). See \cref{app:why_tf_torch} and \cref{app:how_tf_torch} for details. The running of the pipeline is parameterized by the configuration class - \emph{RunnerConfig} (without hyper-parameter tuning) and \emph{HPOConfig} (with hyper-parameter tuning).

\subsection{Why Does {EasyTPP} Support Both TensorFlow and PyTorch} \label{app:why_tf_torch}

TensorFlow and PyTorch are the two most popular Deep Learning (DL) frameworks today. PyTorch has a reputation for being a research-focused framework, and indeed, most of the authors have implemented TPPs in PyTorch, which are used as references by {EasyTPP}. On the other hand, TensorFlow has been widely used in real world applications. For example, Microsoft recommender,\footnote{\url{https://github.com/microsoft/recommenders}.} NVIDIA Merlin\footnote{\url{https://developer.nvidia.com/nvidia-merlin}.} and Alibaba EasyRec\footnote{\url{https://github.com/alibaba/EasyRec}.} are well-known industrial user modeling systems with TensorFlow as the backend. In recent works, TPPs have been introduced to better capture the evolution of the user preference in continuous-time \citep{cont_time_2019,cont_time_rec_2021, ctrec-2019}. To support the use of TPPs by industrial practitioners, we implement an equivalent set of TPPs in TensorFlow. As a result, {EasyTPP} not only helps researchers analyze the strengths and bottlenecks of existing models, but also facilitates the deployment of TPPs in industrial applications.

\subsection{How Does {EasyTPP} Support Both PyTorch and TensorFlow}
\label{app:how_tf_torch}

\begin{figure}
  \begin{center}
    \includegraphics[width=\textwidth]{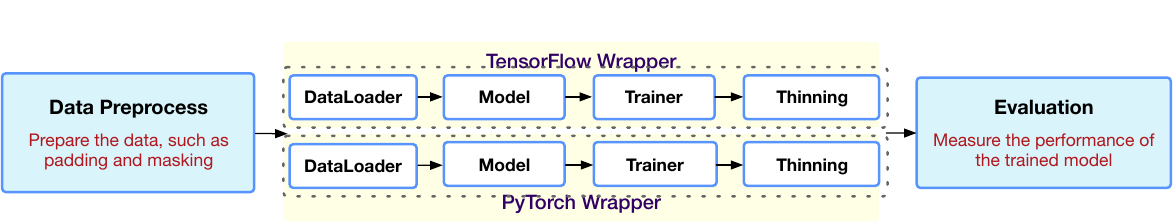}
  \end{center}
  \caption{Illustration of TensorFlow and PyTorch Wrappers in the EasyTPP library.}
  \label{fig:wrapper}
\end{figure}

We implement two equivalent sets of data loaders, models, trainers, thinning samplers in TensorFlow and PyTorch, respectively, then use wrappers to encapsulate them so that they have the same API exposed in the whole training and evaluation pipeline. See \cref{fig:wrapper}.

\subsection{{EasyTPP} for Researchers}

The research groups can inherit from the \emph{BaseModel} to implement their own method in {EasyTPP}. This
opens up a way of standardized and consistent comparisons between different TPPs when exploring new models.

Specifically, if we want to customize a TPP in PyTorch, we need to initialize the model by inheriting the class \emph{TorchBaseModel}:

\begin{lstlisting}[language=Python, caption=Pseudo implementation of customizing a TPP model in PyTorch  using EasyTPP.]

from easy_tpp.model.torch_model.torch_basemodel import TorchBaseModel

# Custom Torch TPP implementations need to
# inherit from the TorchBaseModel interface
class NewModel(TorchBaseModel):
    def __init__(self, model_config):
        super(NewModel, self).__init__(model_config)

    # Forward along the sequence, output the states / intensities at the event times
    def forward(self, batch):
        ...
        return states

    # Compute the loglikelihood loss
    def loglike_loss(self, batch):
        ....
        return loglike

    # Compute the intensities at given sampling times
    # Used in the Thinning sampler
    def compute_intensities_at_sample_times(self, batch, sample_times, **kwargs):
        ...
        return intensities
\end{lstlisting}

Equivalent, if we want to customize a TPP in TensorFlow, we need to initialize the model by inheriting the class \emph{TfBaseModel}:

\begin{lstlisting}[language=Python, caption=Pseudo implementation of customizing a TPP model in TensorFlow using EasyTPP.]

from easy_tpp.model.torch_model.tf_basemodel import TfBaseModel

# Custom Torch TPP implementations need to
# inherit from the TorchBaseModel interface
class NewModel(TfBaseModel):
    def __init__(self, model_config):
        super(NewModel, self).__init__(model_config)

    # Forward along the sequence, output the states / intensities at the event times
    def forward(self, batch):
        ...
        return states

    # Compute the loglikelihood loss
    def loglike_loss(self, batch):
        ....
        return loglike

    # Compute the intensities at given sampling times
    # Used in the Thinning sampler
    def compute_intensities_at_sample_times(self, batch, sample_times, **kwargs):
        ...
        return intensities
\end{lstlisting}

\subsection{{EasyTPP} as a Modeling Library}

A common usage of the package is to train and evaluate some standard TPPs. This can be done by loading black-box-models and data sets from our provided datasets, or by user-defined models and datasets via integration with the defined interfaces.
Listing \ref{fig:library} shows an implementation example of a simple use-case, fitting a TPP model method to a preprocessed dataset from our library.

\begin{lstlisting}[label={fig:library},caption=Example implementation of running a TPP model using EasyTPP.]
import argparse

from easy_tpp.config_factory import Config
from easy_tpp.runner import Runner

def main():
    parser = argparse.ArgumentParser()

    parser.add_argument('--config_dir', 
                        type=str, 
                        required=False, 
                        default='configs/experiment_config.yaml',
                        help='Dir of configuration yaml to train and evaluate the model.')

    parser.add_argument('--experiment_id', 
                        type=str, 
                        required=False, 
                        default='IntensityFree_train',
                        help='Experiment id in the config file.')

    args = parser.parse_args()

    # Build up the configuation for the runner
    config = Config.build_from_yaml_file(args.config_dir, experiment_id=args.experiment_id)

    # Intialize the runner for the pipeline
    model_runner = Runner.build_from_config(config)

    # Start running
    model_runner.run()

if __name__ == '__main__':
    main()
\end{lstlisting}

\section{Model Implementation Details}
\label{app:model_implement}

We have implemented the following TPPs

\begin{itemize}[leftmargin=*]
    \item {\bfseries Multivariate Hawkes Process (MHP)}. We implemented it using Tick~\citep{tick-2017} with an exponential kernel with fixed decays: the base intensity is set to $0.2$ for all dimensions while the decay matrix is set to be a unit matrix plus a small noise. See \href{https://x-datainitiative.github.io/tick/modules/generated/tick.hawkes.HawkesExpKern.html#tick.hawkes.HawkesExpKern}{\textcolor{blue}{online document}} for more details.  
    \item {\bfseries Recurrent marked temporal point process (RMTPP) \textnormal{\citep{du-16-recurrent}}.} We implemented both the Tensorflow and PyTorch version of RMTPP by our own.
    \item {\bfseries Neural Hawkes process (NHP)~\textnormal{\citep{mei-17-neuralhawkes}}} and {\bfseries Attentive neural Hawkes process (AttNHP) \textnormal{\citep{yang-2022-transformer}}.}  The Pytorch implementation mostly comes from the code from  the  public  GitHub  repository at {\small \url{https://github.com/yangalan123/anhp-andtt}}~\citep{yang-2022-transformer} with MIT License. We developed the Tensorflow version of NHP and ttNHP by our own.
    \item {\bfseries Self-attentive Hawkes process (SAHP)~\textnormal{\citep{zhang-2020-self}}} and {\bfseries transformer Hawkes process (THP) \textnormal{\citep{zuo2020transformer}}}. We rewrote the PyTorch versions of SAHP and THP based on the  public  Github repository at {\small \url{https://github.com/yangalan123/anhp-andtt}} \citep{yang-2022-transformer} with MIT License. We developed the Tensorflow versions of the two models by our own.
    \item {\bfseries Intensity-free TPP (IFTPP)~\textnormal{\citep{shchur-20-intensity}}.} 
    The Pytorch implementation mostly comes from the code from  the  public  GitHub  repository at {\small \url{https://github.com/shchur/ifl-tpp}} \citep{shchur-20-intensity} with MIT License. We implemented a Tensorflow version by our own.
    \item {\bfseries Fully network based TPP (FullyNN) \textnormal{\citep{omi-19-fully}}.} We rewrote both the Tensorflow and PyTorch versions of the model faithfully based on the author's code 
    at {\small \url{https://github.com/omitakahiro/NeuralNetworkPointProcess}}. Please not that the model only  considers the number of the types to be one, i.e., the sequence's $K=1$.
    \item {\bfseries ODE-based TPP (ODETPP)~\textnormal{\citep{chen2021neural}}.} We implement a TPP model, in both Tensorflow and PyTorch, with a continuous-time state evolution governed by a neural ODE. It is basically the spatial-temporal point process \citep{chen2021neural} without the spatial component. 
    
\end{itemize}

\subsection{Likelihood Computation Details}
\label{app:loglike}

In this section, we discuss the implementation details of NLL computation in \Cref{eqn:autologlik}.

The integral term in \Cref{eqn:autologlik} is computed using the Monte Carlo approximation given by \citet[Algorithm 1]{mei-17-neuralhawkes}, which samples times $t$.  This yields an unbiased stochastic gradient.  For the number of Monte Carlo samples, we follow the practice of \citet{mei-17-neuralhawkes}: namely, at training time, we match the number of samples to the number of observed events at training time, a reasonable and fast choice, but to estimate log-likelihood when tuning hyperparameters or reporting final results, we take 10 times as many samples. 

At each sampled time $t$, the Monte Carlo method still requires a summation over all events to obtain $\inten{}{t}$.  This summation can be expensive when there are many event types. 
This is not a serious problem for our EasyTPP implementation since it can leverage GPU parallelism.

\subsection{Next Event Prediction}
\label{app:prediction}
It is possible to sample event sequences exactly from any intensity-based model in EasyTPP, using the \defn{thinning algorithm} that is traditionally used for autoregressive point processes \citep{lewis-79-sim,liniger-09-hawkes}.  In general, to apply the thinning algorithm to sample the next event at time $\geq t_0$, it is necessary to have an upper bound on $\{\inten{e}{t}: t \in [t_0,\infty)\}$ for each event type $t$.  An explicit construction for the NHP (or AttNHP) model was given by \citet[Appendix B.3]{mei-17-neuralhawkes}.  

\Cref{sec:bench_process} includes a task-based evaluation where we try to predict the \emph{time} and \emph{type} of just the next event.  More precisely, for each event in each held-out sequence, we attempt to predict its time given only the preceding events, as well as its type given both its true time and the preceding events.

We evaluate the time prediction with average L$_2$ loss (yielding a root-mean-squared error, or \defn{RMSE}) and evaluate the argument prediction with average 0-1 loss (yielding an \defn{error rate}).

Following \citet{mei-17-neuralhawkes}, we use the minimum Bayes risk (MBR) principle to predict the time and type with the lowest expected loss. For completeness, we repeat the general recipe in this section. 

For the $i$-th event, its time $t_{i}$ has density $p_{i}(t) = \inten{}{t} \exp ( -\int_{t_{i-1}}^{t} \inten{}{t'} dt{'} )$.  We choose $\int_{t_{i-1}}^{\infty} t p_{i}(t) dt$ as the time prediction because it has the lowest expected L$_2$ loss. The integral can be estimated using i.i.d.\ samples of $t_{i}$ drawn from $p_{i}(t)$ by the thinning algorithm.

Given the next event time $t_{i}$, we choose the most probable type $\argmax_{e} \inten{e}{t_{i}}$ as the type prediction because it minimizes expected 0-1 loss.

\subsection{Long Horizon Prediction}
\label{app:long_horizon_pred}

The TPP models are typically autoregressive: predicting each future event is conditioned on all the previously predicted events. Following the approach in \citep{xue2022hypro}, we set up a prediction horizon and use OTD to measure the divergence between the ground truth sequence and the predicted sequence within the horizon. For more details about the setup and evaluation protocol, please see Section 5 in \citet{xue2022hypro}.

\section{Dataset Details}
\label{app:dataset}

To comprehensively evaluate the models, we preprocessed one synthetic and five real-world datasets from widely-cited works that contain diverse characteristics
in terms of their application domains and temporal statistics. All preprocessed datasets are available at \href{https://drive.google.com/drive/u/0/folders/1f8k82-NL6KFKuNMsUwozmbzDSFycYvz7}{\textcolor{blue}{Google Drive}}.

\begin{itemize}[leftmargin=*]
    \item {\bfseries Synthetic.} This dataset contains synthetic event sequences from a univariate Hawkes process sampled using {Tick} \citep{tick-2017} whose conditional intensity function is defined by
$$\lambda(t) = \mu +  \sum_{t_{i} < t} \alpha \beta \cdot \exp(-\beta(t - t_{i}))$$
with $\mu = 0.2,  \alpha = 0.8,  \beta = 1.0$. We randomly sampled disjoint train, dev, and test sets with $1200$, $200$ and $400$ sequences.

    \item {\bfseries Amazon}~\textnormal{\citep{amazon-2018}}. This dataset includes time-stamped user product reviews behavior from January, 2008 to October, 2018. 
Each user has a sequence of produce review events with each event containing the timestamp and category of the reviewed product, with each category corresponding to an event type. We work on a subset of $5200$ most active users with an average sequence length of $70$ and then end up with $K=16$ event types.

    \item {\bfseries Retweet~\textnormal{\citep{zhou-2013}}.} This dataset contains time-stamped user retweet event sequences.  The events are categorized into $K = 3$ types: retweets by “small,” “medium” and “large” users. Small users have fewer than $120$ followers, medium users have fewer than $1363$, and the rest are large users. We work on a subset of $5200$ most active users with an average sequence length of $70$.

    \item {\bfseries Taxi~\textnormal{\citep{whong-14-taxi}}.} This dataset tracks the time-stamped taxi pick-up and drop-off events across the five boroughs of the New York City; each (borough, pick-up or drop-off) combination defines an event type, so there are $K=10$ event types in total. We work on a randomly sampled subset of $2000$ drivers and each driver has a sequence.
We randomly sampled disjoint train, dev and test sets with $1400$, $200$ and $400$ sequences.

    \item {\bfseries Taobao}~\textnormal{\citep{xue2022hypro}}. This dataset contains time-stamped user click behaviors on Taobao shopping pages from November 25 to December 03, 2017. Each user has a sequence of item click events with each event containing the timestamp and the category of the item. The categories of all items are first ranked by frequencies and the top $19$ are kept while the rest are merged into one category, with each category corresponding to an event type. We work on a subset of $4800$ most active users with an average sequence length of $150$ and then end up with $K=20$ event types. 

    \item {\bfseries StackOverflow~\textnormal{\citep{snapnets}}.} This dataset has two years of user awards on a question-answering website: each user received a sequence of badges and there are $K=22$ different kinds of badges in total. 
We randomly sampled disjoint train, dev and test sets with $1400,400$ and $400$ sequences from the dataset.
\end{itemize}

\Cref{tab:stats_dataset} shows statistics about each dataset mentioned above.
\begin{table*}[tb]
	\begin{center}
		\begin{small}
			\begin{sc}
				\begin{tabularx}{1.00\textwidth}{l *{1}{S}*{3}{R}*{3}{S}}
					\toprule
					Dataset & \multicolumn{1}{r}{$K$} & \multicolumn{3}{c}{\# of Event Tokens} & \multicolumn{3}{c}{Sequence Length} \\
					\cmidrule(lr){3-8}
					&  & Train & Dev & Test & Min & Mean & Max \\
					\midrule
	Retweet	& $3$ & $369000$ & $62000$ & $61000$ & 10 & $41$ & $97$ \\			
     Taobao & $17$ & $350000$ & $53000$ & $101000$ & $3$ & $51$ & $94$ \\
     Amazon & $16$ & $288000$ & $12000$ & $30000$ & $14$ & $44$ & $94$ \\
					Taxi & $10$ & $51000$ & $7000$ & $14000$ & $36$ & $37$ & $38$ \\
                    StackOverflow & $22$& $90000$& $25000$ & $26000$ & $41$ & $65$ & $101$\\
                    Hawkes-1D & $1$& $55000$& $7000$ & $15000$ & $62$ & $79$ & $95$\\
					\bottomrule
				\end{tabularx}
			\end{sc}
		\end{small}
	\end{center}
	\caption{Statistics of each dataset.}
	\label{tab:stats_dataset}
\end{table*}

\begin{table*}[tb]
	\begin{center}
		\begin{small}
			\begin{sc}
				\begin{tabularx}{1.00\textwidth}{l *{1}{S}*{3}{R}*{3}{S}}
					\toprule
					Model & {Description} & {Value Used} \\
					\midrule
        	& $hidden\_size$ & $32$ \\	
	      	& $time\_emb\_size$ & $16$ \\		
        RMTPP	& $num\_layers$ & $2$ \\	
                    \hdashline
        	& $hidden\_size$ & $64$ \\	
	      	& $time\_emb\_size$ & $16$ \\		
        NHP	    & $ num\_layers$ & $2$ \\	
                    \hdashline
        	& $hidden\_size$ & $32$ \\	
	      	& $time\_emb\_size$ & $16$ \\		
        SAHP	& $ num\_layers$ & $2$ \\	
                & $ num\_heads$ & $2$ \\
                    \hdashline
        	& $hidden\_size$ & $64$ \\	
	      	& $time\_emb\_size$ & $16$ \\		
        THP	    & $ num\_layers$ & $2$ \\
                & $ num\_heads$ & $2$ \\
                    \hdashline
        	& $hidden\_size$ & $32$ \\	
	      	& $time\_emb\_size$ & $16$ \\		
        AttNHP	& $ num\_layers$ & $1$ \\
                & $ num\_heads$ & $2$ \\
                    \hdashline
        	& $hidden\_size$ & $32$ \\	
	ODETPP  & $time\_emb\_size$ & $16$ \\		
        	& $ num\_layers$ & $2$ \\
                    \hdashline
        	& $hidden\_size$ & $32$ \\	
	FullyNN & $time\_emb\_size$ & $16$ \\		
        	& $ num\_layers$ & $2$ \\
                    \hdashline
        	& $hidden\_size$ & $32$ \\	
 IntensityFree  & $time\_emb\_size$ & $16$ \\		
        	& $ num\_layers$ & $2$ \\
					\bottomrule
				\end{tabularx}
			\end{sc}
		\end{small}
	\end{center}
	\caption{Descriptions and values of hyperparameters used for models.}
	\label{tab:stats_hyperparameters}
\end{table*}

\subsection{Padding and Masking}
\label{app:padding}

Given an input sequence $\es{x}{[0, T]} = (t_1, k_1), \ldots, (t_I, k_I)$, for each event, we firstly use an embedding layer to map the event type $k_i$ of each event to a dense vector in higher space; then pass it to the following modules to construct the state embedding $h_i$.

Normally, the input is fed batch-wise into the model; there may exist sequences of events that have unequal-length in the same batch. We pad them to the same length, where the padding mechanism is the same as that in NLP when we pad text sentences.

1. Users can choose to decide whether padding to the beginning or end, padding to max length of batch or a fixed length across the dataset. The function interface and usage are almost the same as those in huggingface/transformers package.

2. The sequence$\_$mask is used to denote whether the event is a padded one or a real one, which will be used in computation of log-likelihood and type/rmse evaluation as well.

A more detailed explanation of dataset preprocessing operation can be found at our \href{https://github.com/ant-research/EasyTemporalPointProcess/blob/main/docs/source/user_guide/dataset.rst#expected-dataset-format-and-data-processing}{\textcolor{blue}{online document}}.

\section{Experiment Details}

\subsection{Setup}
{\bfseries Training Details.} For TPPs, the main hyperparameters to tune are the hidden dimension $D$ of the neural network and the number of layers $L$ of the attention structure (if applicable). In practice, the optimal $D$ for a model was usually $16, 32,64$; the optimal $L$ was usually $1,2,3,4$. To train the parameters for a given generator, we performed early stopping based on log-likelihood on the held-out dev set. The chosen parameters for the main experiments are given in \cref{tab:stats_hyperparameters}.

{\bfseries Computation Cost.} 
All the experiments were conducted on a server with $256$G RAM, a $64$ logical cores CPU (Intel(R) Xeon(R) Platinum 8163 CPU @ 2.50GHz) and one NVIDIA Tesla P100 GPU for acceleration. For training, the batch size is 256 by default. On all the dataset, the training of AttNHP takes most of the time (i.e., around 4 hours) while other models take less than 2 hours.

\subsection{Sanity Check}
\label{app:sanity_checks}

For each model we reproduced in our library, we ran experiments to ensure that our implementation could match the results in the original paper. We used the same hyperparameters as in original papers; we reran each experiment 5 times and took the average. 

In \Cref{tab:main_results_check}, we show the relative differences between the implementations on Retweet and Taxi datasets. As we can see, all the relative differences are within $(-5\%, 5\%)$, indicating that our implementation is close to the original.

\begin{table*}[tbh]
	\begin{center}
		\begin{small}
			\begin{sc}
				\begin{tabularx}{1.00\textwidth}{l *{1}{S}*{2}{R}}
					\toprule
					Model & \multicolumn{2}{c}{Metrics (Time RMSE $/$ Type Error Rate)}  \\
					\cmidrule(lr){2-3}
					& Retweet & Taxi \\
					\midrule	
 RMTPP	& $-4.1\%/-3.5\%$ & $-2.9\%/-3.7\%$\\			
        NHP & $ +3.4\%/+3.1\%$ & $+2.6\%/+3.5\%$ \\
        SAHP & $+1.3\%/+1.7\%$ & $+1.1\%/+1.2\%$\\
        THP & $+1.3\%/+1.8\%$ & $-1.6\%/+1.5\%$ \\
        AttNHP & $+1.2\%/-1.0\%$ & $-1.2\%/-1.2\%$ \\
        ODETPP & $-4.0\%/-3.9\%$ & $-4.3\%/-4.5\%$ \\
        FullyNN & $-5.0\%/$N.A.\phantom{xx} & $-4.1\%/$N.A.\phantom{xx}  \\
        IFTPP & $+3.4\%/+3.1\%$ & $+3.9\%/+3.0\%$ \\
					\bottomrule
				\end{tabularx}
			\end{sc}
		\end{small}
	\end{center}
	\caption{The relative difference between the results of EasyTPP and original implementations.}
	\label{tab:main_results_check}
\end{table*}

\subsection{More Results.} 
\label{app:more_result}


For better visual comparisons, we present the results in \cref{fig:main_results_loglike}, \cref{fig:main_results_retweet_lt_generation} and \cref{fig:main_results_taxi_lt_generation} also in the form of tables, see \cref{tab:main_results_numbers_loglike} and \cref{tab:main_results_retweet_lt_generation}.

The relative difference between the results of Torch and TensorFlow implementations can be found in \cref{tab:main_results_rel_diff}.

\begin{table*}[tbh]
	\begin{center}
		\begin{small}
			\begin{sc}
				\begin{tabularx}{1.00\textwidth}{l *{1}{S}*{4}{R}}
					\toprule
					Model & \multicolumn{4}{c}{OTD}  \\
					\cmidrule(lr){2-5}
					& Retweet  & Retweet& Taxi  & Taxi  \\
     & avg $5$ events & avg $10$ events & avg $5$ events & avg $10$ events\\
					\midrule
	MHP &  $5.128$  & $11.270$  & $4.633$  & $12.784$ \\
            &  $(0.040)$  & $(0.091)$  & $(0.037)$  & $(0.111)$ \\
            \midrule
        RMTPP	& $5.107$ & $10.255$ & $4.401$ & $12.045$  \\	
        &  $(0.041)$  & $(0.099)$  & $(0.030)$  & $(0.114)$ \\
            \midrule
        NHP & $5.080$ & $10.470$ & $4.412$ & $12.110$\\
        &  $(0.042)$  & $(0.085)$  & $(0.032)$  & $(0.125)$ \\
            \midrule
        SAHP & $5.092$ & $10.475$ & $4.422$ & $12.051$ \\
        &  $(0.039)$  & $(0.079)$  & $(0.039)$  & $(0.139)$ \\
            \midrule
        THP & $5.091$ & $\underline{10.450}$ & $\underline{4.398}$ & $\underline{11.875}$ \\
        &  $(0.052)$  & $(0.090)$  & $(0.041)$  & $(0.108)$ \\
            \midrule
        AttNHP & $\underline{5.077}$ & $10.447$ & $4.420$ & $12.102$ \\
        &  $(0.039)$  & $(0.090)$  & $(0.044)$  & $(0.109)$ \\
            \midrule
        ODETPP & $5.115$ & $10.483$ & $4.408$ & $12.095$  \\
        &  $(0.041)$  & $(0.088)$  & $(0.039)$  & $(0.100)$ \\
            \midrule
        FullyNN & N.A. & N.A. & N.A. & N.A.  \\
        & N.A. & N.A. & N.A. & N.A.\\
            \midrule
        IFTPP & $5.079$ & $10.513$ & $4.501$ & $12.052$  \\
        &  $(0.061)$  & $(0.115)$  & $(0.032)$  & $(0.121)$ \\
					\bottomrule
				\end{tabularx}
			\end{sc}
		\end{small}
	\end{center}
	\caption{Long horizon prediction on Retweet and Taxi data (the first row is the prediction value while the second row is the standard deviation).}
	\label{tab:main_results_retweet_lt_generation}
\end{table*}

\begin{table*}[tbh]
	\begin{center}
		\begin{footnotesize}
			\begin{sc}
				\begin{tabularx}{1.00\textwidth}{l *{1}{S}*{5}{R}}
					\toprule
					Model & \multicolumn{5}{c}{Rel Diff on Time RMSE (1st Row) and Type Error Rate (2nd Row)}  \\
					\cmidrule(lr){2-6}
					& Amazon & Retweet & Taxi & Taobao & StackOverflow   \\
					\midrule
 	RMTPP	& $-0.2\%$ & $+1.0\%$ & $+0.1\%$ & $+0.1\%$ & $+0.4\%$  \\
  		& $+0.5\%$ & $+1.3\%$ & $+0.6\%$ & $+0.2\%$ & $-0.7\%$ \\
        \midrule
        NHP & $+0.7\%$ & $+0.5\%$ & $-0.2\%$ & $+0.1\%$ & $-0.1\%$ \\
         & $+0.6\%$ & $+1.4\%$ & $+0.4\%$ & $-0.3\%$ & $-0.1\%$ \\
        \midrule
        SAHP & $-0.8\%$ & $+0.7\%$ & $-0.8\%$ & $+0.4\%$ & $0.3\%$ \\
         & $+0.6\%$ & $+0.6\%$ & $-0.6\%$ & $+0.4\%$ & $0.3\%$ \\
         \midrule
        THP & $+0.6\%$ & $+0.6\%$ & $-0.2\%$ & $-0.5\%$ & $0.6\%$ \\
         & $+1.2\%$ & $+0.9\%$ & $-0.6\%$ & $+0.7\%$ & $0.4\%$ \\
        \midrule
        AttNHP & $+0.4\%$ & $+0.4\%$ & $+0.3\%$ & $-0.1\%$ & $-0.2\%$ \\
         & $+0.2\%$ & $-0.7\%$ & $-0.6\%$ & $+0.4\%$ & $+0.2\%$ \\
        \midrule
        ODETPP & $-0.5\%$ & $+1.1\%$ & $+0.9\%$ & $+0.6\%$ & $0.4\%$ \\
         & $+0.8\%$ & $+1.3\%$ & $+1.1\%$ & $-0.5\%$ & $-0.5\%$ \\
        \midrule
        FullyNN & $+0.5\%$ & $-0.7\%$ & $-0.3\%$ & $-0.3\%$ & $+0.2\%$\\

        & NA&NA&NA&NA&NA\\
        \midrule
        IFTPP & $-0.9\%$ & $+1.0\%$ & $+0.4\%$ & $+0.6\%$ & $+0.3\%$ \\
         & $+0.4\%$ & $-0.7\%$ & $-0.3\%$ & $+0.2\%$ & $+0.2\%$ \\
					\bottomrule
				\end{tabularx}
			\end{sc}
		\end{footnotesize}
	\end{center}
	\caption{Relative difference between Torch and TensorFlow implementations of methods in \cref{tab:main_results_pred_main_text}.}
    \label{tab:main_results_rel_diff}
\end{table*}

\begin{table*}[tbh]
	\begin{center}
		\begin{small}
			\begin{sc}
				\begin{tabularx}{1.00\textwidth}{l *{1}{S}*{3}{R}}
					\toprule
					Model & \multicolumn{3}{c}{Metrics (Loglike)}  \\
					\cmidrule(lr){2-4}
					& Synthetic & Retweet & Taxi    \\
					\midrule
MHP	& $-3.150$ & $-5.949$ & $-1.466$  \\	
 & $(0.028)$ & $(0.033)$ & $(0.011)$  \\		\midrule
 RMTPP	& $-0.998$ & $-4.237$ & $-0.227$   \\	
 & $(0.009)$ & $(0.033)$ & $(0.001)$  \\
 \midrule
        NHP & $-0.443$ & $-4.137$ & $-0.208$ \\
        & $(0.004)$ & $(0.050)$ & $(0.001)$  \\
        \midrule
        SAHP & $-1.337$ & $-5.009$ & $-0.478$ \\
        & $(0.013)$ & $(0.044)$ & $(0.003)$  \\
        \midrule
        THP & $-1.238$ & $-4.560$ & $-0.442$  \\
        & $(0.011)$ & $(0.041)$ & $(0.004)$  \\
        \midrule
        AttNHP & $-1.001$ & $-4.756$ & $-0.491$  \\
        & $(0.008)$ & $(0.052)$ & $(0.004)$  \\
        \midrule
        ODETPP & $-1.007$ & $-4.527$ & $-0.217$ \\
        & $(0.007)$ & $(0.044)$ & $(0.002)$  \\
        \midrule
        FullyNN & $-2.318$ & $-5.889$ & $-1.317$ \\
        & $(0.014)$ & $(0.032)$ & $(0.009)$  \\
        \midrule
        IFTPP & $\underline{0.186}$ & $\underline{-0.212}$ & $\underline{0.019}$  \\
        & $(0.002)$ & $(0.002)$ & $(0.001)$   \\
					\bottomrule
				\end{tabularx}
			\end{sc}
		\end{small}
	\end{center}
	\caption{Performance in numbers of all methods in} \cref{fig:main_results_loglike} (the first row is the prediction value while the
second row is the standard deviation).
	\label{tab:main_results_numbers_loglike}
\end{table*}

\section{Additional Note}

\subsection{Citation Count in ArXiv}
\label{app:cite}
We search the TPP-related articles in ArXiv {\small \url{https://arxiv.org/}} using their own search engine in three folds:
\begin{itemize}[leftmargin=*]
    \item Temporal point process: we search through the abstract of articles which contains the term `temporal point process'.
    \item Hawkes process: we search through the abstract of articles with  the term `hawkes process' but without the term `temporal point process'.
    \item Temporal event sequence: we search through the abstract of articles which include the term `temporal event sequence' but exclude the term `hawkes process' and `temporal point process'.
\end{itemize}

We group the articles found out by the search engine by years and report it in \cref{fig:arxiv}.

\end{document}

%% file: math_commands.tex
\usepackage{amsmath,amsfonts,bm}

\def\eqref#1{equation~\ref{#1}}

\def\1{\bm{1}}

\DeclareMathAlphabet{\mathsfit}{\encodingdefault}{\sfdefault}{m}{sl}
\SetMathAlphabet{\mathsfit}{bold}{\encodingdefault}{\sfdefault}{bx}{n}

\DeclareMathOperator*{\argmax}{arg\,max}

%% file: main.bbl
\begin{thebibliography}{48}
\providecommand{\natexlab}[1]{#1}
\providecommand{\url}[1]{\texttt{#1}}
\expandafter\ifx\csname urlstyle\endcsname\relax
  \providecommand{\doi}[1]{doi: #1}\else
  \providecommand{\doi}{doi: \begingroup \urlstyle{rm}\Url}\fi

\bibitem[Abadi et~al.(2016)Abadi, Barham, Chen, Chen, Davis, Dean, Devin, Ghemawat, Irving, Isard, et~al.]{abadi2016tensorflow}
Abadi, M., Barham, P., Chen, J., Chen, Z., Davis, A., Dean, J., Devin, M., Ghemawat, S., Irving, G., Isard, M., et~al.
\newblock Tensorflow: a system for large-scale machine learning.
\newblock In \emph{OSDI'16}, volume~16, pp.\  265--283, 2016.

\bibitem[Akiba et~al.(2019)Akiba, Sano, Yanase, Ohta, and Koyama]{optuna_2019}
Akiba, T., Sano, S., Yanase, T., Ohta, T., and Koyama, M.
\newblock Optuna: A next-generation hyperparameter optimization framework.
\newblock In \emph{Proceedings of the 25rd {ACM} {SIGKDD} International Conference on Knowledge Discovery and Data Mining}, 2019.

\bibitem[{Bacry} et~al.(2017){Bacry}, {Bompaire}, {Ga{\"i}ffas}, and {Poulsen}]{tick-2017}
{Bacry}, E., {Bompaire}, M., {Ga{\"i}ffas}, S., and {Poulsen}, S.
\newblock \href {http://arxiv.org/abs/1707.03003} {{tick: a Python library for statistical learning, with a particular emphasis on time-dependent modeling}}.
\newblock \emph{ArXiv e-prints}, 2017.

\bibitem[Bae et~al.(2023)Bae, Ahmed, Tung, and Oliveira]{bae2023meta}
Bae, W., Ahmed, M.~O., Tung, F., and Oliveira, G.~L.
\newblock \href {https://arxiv.org/abs/2301.12023} {Meta temporal point processes}.
\newblock In \emph{Proceedings of the International Conference on Learning Representations (ICLR)}, 2023.

\bibitem[Bai et~al.(2019)Bai, Zou, Zhao, Du, Liu, Nie, and Wen]{ctrec-2019}
Bai, T., Zou, L., Zhao, W.~X., Du, P., Liu, W., Nie, J.-Y., and Wen, J.-R.
\newblock \href {https://doi.org/10.1145/3331184.3331199} {Ctrec: A long-short demands evolution model for continuous-time recommendation}.
\newblock In \emph{Proceedings of the 42nd International ACM SIGIR Conference on Research and Development in Information Retrieval}, SIGIR'19, pp.\  675–684, New York, NY, USA, 2019. Association for Computing Machinery.
\newblock ISBN 9781450361729.

\bibitem[Bao \& Zhang(2021)Bao and Zhang]{cont_time_2019}
Bao, J. and Zhang, Y.
\newblock \href {https://doi.org/10.1145/3459637.3482202} {Time-aware recommender system via continuous-time modeling}.
\newblock In \emph{Proceedings of the 30th ACM International Conference on Information \& Knowledge Management}, CIKM '21, pp.\  2872–2876, New York, NY, USA, 2021. Association for Computing Machinery.
\newblock ISBN 9781450384469.

\bibitem[Boyd et~al.(2020)Boyd, Bamler, Mandt, and Smyth]{boyd-20-vae}
Boyd, A., Bamler, R., Mandt, S., and Smyth, P.
\newblock \href {https://arxiv.org/pdf/2011.03231.pdf} {User-dependent neural sequence models for continuous-time event data}.
\newblock In \emph{Advances in Neural Information Processing Systems (NeurIPS)}, 2020.

\bibitem[Brown et~al.(2020)Brown, Mann, Ryder, Subbiah, Kaplan, Dhariwal, Neelakantan, Shyam, Sastry, Askell, et~al.]{brown-2020-gpt}
Brown, T.~B., Mann, B., Ryder, N., Subbiah, M., Kaplan, J., Dhariwal, P., Neelakantan, A., Shyam, P., Sastry, G., Askell, A., et~al.
\newblock \href {https://papers.nips.cc/paper/2020/file/1457c0d6bfcb4967418bfb8ac142f64a-Paper.pdf} {Language models are few-shot learners}.
\newblock In \emph{Advances in Neural Information Processing Systems (NeurIPS)}, 2020.

\bibitem[Chen et~al.(2021)Chen, Amos, and Nickel]{chen2021neural}
Chen, R.~T., Amos, B., and Nickel, M.
\newblock Neural spatio-temporal point processes.
\newblock \emph{ICLR}, 2021.

\bibitem[Cram{\'e}r(1969)]{cramer1969historical}
Cram{\'e}r, H.
\newblock Historical review of {F}ilip {L}undberg's works on risk theory.
\newblock \emph{Scandinavian Actuarial Journal}, 1969\penalty0 (sup3):\penalty0 6--12, 1969.

\bibitem[Daley \& Vere-Jones(2007)Daley and Vere-Jones]{daley-07-poisson}
Daley, D.~J. and Vere-Jones, D.
\newblock \href {https://link.springer.com/book/10.1007/978-0-387-49835-5} {\emph{An Introduction to the Theory of Point Processes, Volume {II}: General Theory and Structure}}.
\newblock Springer, 2007.

\bibitem[Du et~al.(2016)Du, Dai, Trivedi, Upadhyay, Gomez-Rodriguez, and Song]{du-16-recurrent}
Du, N., Dai, H., Trivedi, R., Upadhyay, U., Gomez-Rodriguez, M., and Song, L.
\newblock \href {https://www.kdd.org/kdd2016/papers/files/rpp1081-duA.pdf} {Recurrent marked temporal point processes: Embedding event history to vector}.
\newblock In \emph{Proceedings of the ACM SIGKDD International Conference on Knowledge Discovery and Data Mining}, 2016.

\bibitem[Fan et~al.(2021)Fan, Liu, Zhang, Xiong, Zheng, and Yu]{cont_time_rec_2021}
Fan, Z., Liu, Z., Zhang, J., Xiong, Y., Zheng, L., and Yu, P.~S.
\newblock \href {https://doi.org/10.1145/3459637.3482242} {Continuous-time sequential recommendation with temporal graph collaborative transformer}.
\newblock In \emph{Proceedings of the 30th ACM International Conference on Information \& Knowledge Management}, CIKM '21, pp.\  433–442, New York, NY, USA, 2021. Association for Computing Machinery.
\newblock ISBN 9781450384469.

\bibitem[Hasbrouck(1991)]{hasbrouck-1991}
Hasbrouck, J.
\newblock \href {https://doi.org/https://doi.org/10.1111/j.1540-6261.1991.tb03749.x} {Measuring the information content of stock trades}.
\newblock \emph{The Journal of Finance}, 46\penalty0 (1):\penalty0 179--207, 1991.

\bibitem[Hawkes(1971)]{hawkes-71}
Hawkes, A.~G.
\newblock \href {https://pdfs.semanticscholar.org/c082/06b44dd1f0ea54bd073e4effaf2e4483169b.pdf} {Spectra of some self-exciting and mutually exciting point processes}.
\newblock \emph{Biometrika}, 1971.

\bibitem[Hernandez et~al.(2017)Hernandez, {\'A}lvarez, Fabra, and Ezpeleta]{hernandez2017analysis}
Hernandez, S., {\'A}lvarez, P., Fabra, J., and Ezpeleta, J.
\newblock Analysis of users’ behavior in structured e-commerce websites.
\newblock \emph{IEEE Access}, 5:\penalty0 11941--11958, 2017.

\bibitem[Jin et~al.(2020)Jin, Guo, Chen, Weiskopf, Gotz, and Cao]{jin2020visual}
Jin, Z., Guo, S., Chen, N., Weiskopf, D., Gotz, D., and Cao, N.
\newblock Visual causality analysis of event sequence data.
\newblock \emph{IEEE transactions on visualization and computer graphics}, 27\penalty0 (2):\penalty0 1343--1352, 2020.

\bibitem[Ke~Zhou \& Song(2013)Ke~Zhou and Song]{zhou-2013}
Ke~Zhou, H.~Z. and Song, L.
\newblock Learning triggering kernels for multi-dimensional hawkes processes.
\newblock In \emph{Proceedings of the International Conference on Machine Learning (ICML)}, 2013.

\bibitem[Kingma \& Ba(2015)Kingma and Ba]{kingma-15}
Kingma, D. and Ba, J.
\newblock \href {https://arxiv.org/pdf/1412.6980.pdf} {{Adam}: A method for stochastic optimization}.
\newblock In \emph{Proceedings of the International Conference on Learning Representations (ICLR)}, 2015.

\bibitem[Leskovec \& Krevl(2014)Leskovec and Krevl]{snapnets}
Leskovec, J. and Krevl, A.
\newblock \href {https://snap.stanford.edu/data/} {{SNAP} {D}atasets: Stanford large network dataset collection}, 2014.

\bibitem[Lewis \& Shedler(1979)Lewis and Shedler]{lewis-79-sim}
Lewis, P.~A. and Shedler, G.~S.
\newblock \href {https://onlinelibrary.wiley.com/doi/10.1002/nav.3800260304} {Simulation of nonhomogeneous {Poisson} processes by thinning}.
\newblock \emph{Naval Research Logistics Quarterly}, 1979.

\bibitem[Liniger(2009)]{liniger-09-hawkes}
Liniger, T.~J.
\newblock \href {https://www.research-collection.ethz.ch/bitstream/handle/20.500.11850/151886/eth-1112-02.pdf} {\emph{Multivariate {Hawkes} processes}}.
\newblock Diss., Eidgen{\"o}ssische Technische Hochschule ETH Z{\"u}rich, Nr. 18403, 2009.

\bibitem[Mei \& Eisner(2017)Mei and Eisner]{mei-17-neuralhawkes}
Mei, H. and Eisner, J.
\newblock \href {https://arxiv.org/abs/1612.09328} {The neural {H}awkes process: {A} neurally self-modulating multivariate point process}.
\newblock In \emph{Advances in Neural Information Processing Systems (NeurIPS)}, 2017.

\bibitem[Mei et~al.(2019)Mei, Qin, and Eisner]{mei-19-smoothing}
Mei, H., Qin, G., and Eisner, J.
\newblock \href {https://arxiv.org/pdf/1905.05570.pdf} {Imputing missing events in continuous-time event streams}.
\newblock In \emph{Proceedings of the International Conference on Machine Learning (ICML)}, 2019.

\bibitem[Mei et~al.(2020)Mei, Qin, Xu, and Eisner]{mei-2020-datalog}
Mei, H., Qin, G., Xu, M., and Eisner, J.
\newblock \href {https://www.cs.jhu.edu/~jason/papers/#mei-et-al-2020-icml} {Neural {D}atalog through time: Informed temporal modeling via logical specification}.
\newblock In \emph{Proceedings of the International Conference on Machine Learning (ICML)}, 2020.

\bibitem[Nakano et~al.(2021)Nakano, Hilton, Balaji, Wu, Ouyang, Kim, Hesse, Jain, Kosaraju, Saunders, et~al.]{nakano2021webgpt}
Nakano, R., Hilton, J., Balaji, S., Wu, J., Ouyang, L., Kim, C., Hesse, C., Jain, S., Kosaraju, V., Saunders, W., et~al.
\newblock \href {https://arxiv.org/abs/2112.09332} {Webgpt: Browser-assisted question-answering with human feedback}.
\newblock \emph{arXiv preprint arXiv:2112.09332}, 2021.

\bibitem[Ni(2018)]{amazon-2018}
Ni, J.
\newblock \href {https://nijianmo.github.io/amazon/} {Amazon review data}, 2018.

\bibitem[Ogata(1988)]{ogata1988statistical}
Ogata, Y.
\newblock Statistical models for earthquake occurrences and residual analysis for point processes.
\newblock \emph{J. Am. Stat. Assoc}, 83\penalty0 (401):\penalty0 9--27, 1988.

\bibitem[Omi et~al.(2019)Omi, Ueda, and Aihara]{omi-19-fully}
Omi, T., Ueda, N., and Aihara, K.
\newblock \href {https://arxiv.org/abs/1905.09690} {Fully neural network based model for general temporal point processes}.
\newblock In \emph{Advances in Neural Information Processing Systems (NeurIPS)}, 2019.

\bibitem[OpenAI(2023)]{gpt4}
OpenAI.
\newblock \href {https://arxiv.org/abs/2303.08774.pdf} {{GPT}-4 technical report}.
\newblock \emph{arXiv preprint arXiv:2303.08774}, 2023.

\bibitem[Paszke et~al.(2019)Paszke, Gross, Massa, Lerer, Bradbury, Chanan, Killeen, Lin, Gimelshein, Antiga, et~al.]{paszke2019pytorch}
Paszke, A., Gross, S., Massa, F., Lerer, A., Bradbury, J., Chanan, G., Killeen, T., Lin, Z., Gimelshein, N., Antiga, L., et~al.
\newblock Pytorch: An imperative style, high-performance deep learning library.
\newblock \emph{arXiv preprint arXiv:1912.01703}, 2019.

\bibitem[Qu et~al.(2023)Qu, Tan, Xue, Shi, Zhang, and Mei]{qu-2022-rltpp}
Qu, C., Tan, X., Xue, S., Shi, X., Zhang, J., and Mei, H.
\newblock \href {https://arxiv.org/abs/2201.12569} {Bellman meets hawkes: Model-based reinforcement learning via temporal point processes}.
\newblock In \emph{Proceedings of the AAAI Conference on Artificial Intelligence}, 2023.

\bibitem[Rubanova et~al.(2019)Rubanova, Chen, and Duvenaud]{rubanova2019latent}
Rubanova, Y., Chen, R.~T., and Duvenaud, D.~K.
\newblock Latent ordinary differential equations for irregularly-sampled time series.
\newblock \emph{Advances in Neural Information Processing Systems (NeurIPS)}, 32, 2019.

\bibitem[Shchur et~al.(2020)Shchur, Bilo{\v{s}}, and G{\"u}nnemann]{shchur-20-intensity}
Shchur, O., Bilo{\v{s}}, M., and G{\"u}nnemann, S.
\newblock \href {https://arxiv.org/abs/1909.12127} {Intensity-free learning of temporal point processes}.
\newblock In \emph{Proceedings of the International Conference on Learning Representations (ICLR)}, 2020.

\bibitem[Shi et~al.(2023)Shi, Xue, Wang, Zhou, Zhang, Zhou, Tan, and Mei]{shi2023language}
Shi, X., Xue, S., Wang, K., Zhou, F., Zhang, J.~Y., Zhou, J., Tan, C., and Mei, H.
\newblock \href {https://arxiv.org/abs/2305.16646} {Language models can improve event prediction by few-shot abductive reasoning}.
\newblock \emph{arXiv preprint arXiv:2305.16646}, 2023.

\bibitem[Vaswani et~al.(2017)Vaswani, Shazeer, Parmar, Uszkoreit, Jones, Gomez, Kaiser, and Polosukhin]{vaswani-2017-transformer}
Vaswani, A., Shazeer, N., Parmar, N., Uszkoreit, J., Jones, L., Gomez, A.~N., Kaiser, L., and Polosukhin, I.
\newblock \href {https://arxiv.org/pdf/1706.03762.pdf} {Attention is all you need}.
\newblock In \emph{Advances in Neural Information Processing Systems (NeurIPS)}, 2017.

\bibitem[Wang et~al.(2019)Wang, Singh, Michael, Hill, Levy, and Bowman]{glue-2019}
Wang, A., Singh, A., Michael, J., Hill, F., Levy, O., and Bowman, S.~R.
\newblock \href {http://arxiv.org/abs/1804.07461} {{GLUE:} {A} multi-task benchmark and analysis platform for natural language understanding}.
\newblock \emph{ICLR}, 2019.

\bibitem[Wen et~al.(2023)Wen, Zhou, Zhang, Chen, Ma, Yan, and Sun]{wen2022transformers}
Wen, Q., Zhou, T., Zhang, C., Chen, W., Ma, Z., Yan, J., and Sun, L.
\newblock Transformers in time series: A survey.
\newblock In \emph{International Joint Conference on Artificial Intelligence(IJCAI)}, 2023.

\bibitem[Whong(2014)]{whong-14-taxi}
Whong, C.
\newblock \href {https://chriswhong.com/open-data/foil_nyc_taxi/} {F{OIL}ing {NYC}’s taxi trip data}, 2014.

\bibitem[Williams et~al.(2020)Williams, Degleris, Wang, and Linderman]{williams2020point}
Williams, A., Degleris, A., Wang, Y., and Linderman, S.
\newblock Point process models for sequence detection in high-dimensional neural spike trains.
\newblock \emph{Advances in neural information processing systems}, 33:\penalty0 14350--14361, 2020.

\bibitem[Wolf et~al.(2020)Wolf, Debut, Sanh, Chaumond, Delangue, Moi, Cistac, Rault, Louf, Funtowicz, Davison, Shleifer, von Platen, Ma, Jernite, Plu, Xu, Scao, Gugger, Drame, Lhoest, and Rush]{wolf-etal-2020-transformers}
Wolf, T., Debut, L., Sanh, V., Chaumond, J., Delangue, C., Moi, A., Cistac, P., Rault, T., Louf, R., Funtowicz, M., Davison, J., Shleifer, S., von Platen, P., Ma, C., Jernite, Y., Plu, J., Xu, C., Scao, T.~L., Gugger, S., Drame, M., Lhoest, Q., and Rush, A.~M.
\newblock \href {https://www.aclweb.org/anthology/2020.emnlp-demos.6} {Transformers: State-of-the-art natural language processing}.
\newblock In \emph{Proceedings of the 2020 Conference on Empirical Methods in Natural Language Processing: System Demonstrations}, pp.\  38--45, Online, October 2020. Association for Computational Linguistics.

\bibitem[Xiao et~al.(2017)Xiao, Yan, Yang, Zha, and Chu]{xiao-17-modeling}
Xiao, S., Yan, J., Yang, X., Zha, H., and Chu, S.
\newblock \href {https://arxiv.org/abs/1705.08982} {Modeling the intensity function of point process via recurrent neural networks}.
\newblock In \emph{Proceedings of the AAAI Conference on Artificial Intelligence}, 2017.

\bibitem[Xu(2018)]{xu2018poppy}
Xu, H.
\newblock Poppy: A point process toolbox based on pytorch.
\newblock \emph{arXiv preprint arXiv:1810.10122}, 2018.

\bibitem[Xue et~al.(2022)Xue, Shi, Zhang, and Mei]{xue2022hypro}
Xue, S., Shi, X., Zhang, Y.~J., and Mei, H.
\newblock \href {https://arxiv.org/abs/2210.01753} {Hypro: A hybridly normalized probabilistic model for long-horizon prediction of event sequences}.
\newblock In \emph{Advances in Neural Information Processing Systems (NeurIPS)}, 2022.

\bibitem[Yang et~al.(2022)Yang, Mei, and Eisner]{yang-2022-transformer}
Yang, C., Mei, H., and Eisner, J.
\newblock \href {https://arxiv.org/abs/2201.00044} {Transformer embeddings of irregularly spaced events and their participants}.
\newblock In \emph{Proceedings of the International Conference on Learning Representations (ICLR)}, 2022.

\bibitem[Zhang et~al.(2020)Zhang, Lipani, Kirnap, and Yilmaz]{zhang-2020-self}
Zhang, Q., Lipani, A., Kirnap, O., and Yilmaz, E.
\newblock \href {http://proceedings.mlr.press/v119/zhang20q/zhang20q.pdf} {Self-attentive {H}awkes process}.
\newblock In \emph{Proceedings of the International Conference on Machine Learning (ICML)}, 2020.

\bibitem[Zhu et~al.(2021)Zhu, Liu, Yang, Zhang, and He]{fuxi-2021}
Zhu, J., Liu, J., Yang, S., Zhang, Q., and He, X.
\newblock \href {https://doi.org/10.1145/3459637.3482486} {Open benchmarking for click-through rate prediction}.
\newblock In Demartini, G., Zuccon, G., Culpepper, J.~S., Huang, Z., and Tong, H. (eds.), \emph{{CIKM} '21: The 30th {ACM} International Conference on Information and Knowledge Management, Virtual Event, Queensland, Australia, November 1 - 5, 2021}, pp.\  2759--2769. {ACM}, 2021.

\bibitem[Zuo et~al.(2020)Zuo, Jiang, Li, Zhao, and Zha]{zuo2020transformer}
Zuo, S., Jiang, H., Li, Z., Zhao, T., and Zha, H.
\newblock \href {https://arxiv.org/pdf/2002.09291.pdf} {Transformer {H}awkes process}.
\newblock In \emph{International Conference on Machine Learning}, pp.\  11692--11702. PMLR, 2020.

\end{thebibliography}
